\documentclass[3p,times]{elsarticle}

\usepackage{amssymb}
\usepackage{amsthm}
\usepackage{booktabs}
\usepackage{multirow}
\usepackage{graphicx}
\usepackage{lscape}
\usepackage{adjustbox}
\usepackage{mathtools}
\usepackage{rotating}
\usepackage{caption}
\usepackage[aboveskip=1.5pt]{subcaption}
\usepackage{hyperref}

\begin{document}

\begin{frontmatter}


\title{A comprehensive analysis of concept drift locality in data streams}


\author[inst1]{Gabriel J. Aguiar}
\ead{aguiargj@vcu.edu}

\author[inst1]{Alberto Cano}
\ead{acano@vcu.edu}

\affiliation[inst1]{organization={Department of Computer Science, Virginia Commonwealth University},
            city={\\Richmond},
            state={Virginia},
            country={USA}}

\begin{abstract}
Adapting to drifting data streams is a significant challenge in online learning. Concept drift must be detected for effective model adaptation to evolving data properties. Concept drift can impact the data distribution entirely or partially, which makes it difficult for drift detectors to accurately identify the concept drift. Despite the numerous concept drift detectors in the literature, standardized procedures and benchmarks for comprehensive evaluation considering the locality of the drift are lacking. We present a novel categorization of concept drift based on its locality and scale. A systematic approach leads to a set of 2,760 benchmark problems, reflecting various difficulty levels following our proposed categorization. We conduct a comparative assessment of 9 state-of-the-art drift detectors across diverse difficulties, highlighting their strengths and weaknesses for future research. We examine how drift locality influences the classifier performance and propose strategies for different drift categories to minimize the recovery time. Lastly, we provide lessons learned and recommendations for future concept drift research. Our benchmark data streams and experiments are publicly available at \href{https://github.com/gabrieljaguiar/locality-concept-drift}{https://github.com/gabrieljaguiar/locality-concept-drift}.
\end{abstract}


\end{frontmatter}

\section{Introduction}

Modern data sources continuously generate information characterized by both volume and velocity, flooding learning systems with a constant flow of data. This scenario is commonly referred to as data streams~\cite{gama2004learning, bahri2021data}. Traditional classification methods, designed for static data, struggle to keep up with the ever-changing characteristics of these incoming instances~\cite{gama2004learning, gama2010knowledge}. Given the dynamic nature of data streams, it becomes essential for learning methods to adapt and acquire knowledge about emerging concepts over time. This phenomenon is known as concept drift~\cite{gama2014survey}, and it can manifest in various ways, including shifts in class distribution and decision boundaries~\cite{aguiar2022survey}, and the emergence of new features or classes~\cite{korycki2021concept}. If not detected and addressed effectively, concept drift can significantly degrade predictive performance, as knowledge learned from older concepts may not be useful anymore to classify recent instances~\cite{viniski2021case}.

In recent years, the issue of concept drift has garnered significant attention within the research community across various domains, including sensors, robotics, system monitoring, and anomaly detection~\cite{suarez2022survey}. Current research in this field is tackling increasingly complex challenges. These challenges include accurately detecting concept drift within unstructured and noisy datasets~\cite{lu2016concept}, providing understandable explanations for concept drift~\cite{liu2017regional}, and effectively responding to drift by adapting relevant knowledge~\cite{barros2017rddm}. When we extend these concerns to scenarios involving multiple classes, we encounter a complex and perplexing scenario that actually occurs in many real-life applications. Detecting concept drift in such contexts becomes exceptionally demanding, as we must account for the evolving nature of multiple classes~\cite{korycki2021concept, gulcan2023unsupervised}. In addition to the challenges previously mentioned, it is important to note that the location of concept drift within the feature space significantly influences both the performance of classifiers and the effectiveness of drift detection methods~\cite{korycki2021concept, gama2006learning}. However, there is a lack of studies that evaluate drift detectors under varying degrees of drift locality or provide benchmark datasets to support research in this crucial area.

\noindent \textbf{Motivation.} While the literature offers numerous concept drift detectors, there remains a notable absence of standardized procedures and benchmarks for a comprehensive assessment of these methods when considering the locality of concept drifts. Specifically, there is a lack of dedicated benchmarks suitable for evaluating drift detectors across a diverse spectrum of challenges, particularly those tied to the locality of concept drift. An in-depth experimental comparison of state-of-the-art drift detectors, applied to a diverse set of challenges, would provide valuable insights into the performance of these detectors under various conditions. Moreover, existing studies often focus on specific subsets of detectors and data challenges, typically limited to binary class data. These studies often fail to provide insights into which aspects of concept drift should be considered. Therefore, we propose a comprehensive study to understand and assess the performance of drift detectors across a wide array of difficulties and analyze how the locality of a concept drift impacts its detection.

\noindent \textbf{Overview and main contributions.} This paper presents a comprehensive study for benchmarking and evaluating the impact of the locality and magnitudes of a concept drift on the classifier or drift detector. We systematically identify critical challenges within this domain and leverage them to create a set of benchmark problems that encompass various difficulties, guided by a novel concept drift categorization. Furthermore, we conduct a comparative evaluation of nine state-of-the-art drift detectors across this wide range of difficulty. This analysis not only identifies the top-performing detectors but also sheds light on their specific strengths and weaknesses, providing valuable insights for future research in drift detection. The main contributions of this paper can be summarized as follows:

\begin{itemize}
    \item \textbf{Concept drift locality categorization.} We present a novel categorization that organizes concept drift based on the number of affected classes and the magnitude of the change. This categorization empowers us and future researchers to evaluate the effect of various concept drift levels when proposing new drift detectors or classifiers. This categorization guides the creation of benchmark problems, ranging from scenarios affecting only one class to those that transform the entire data stream. The comprehensive set of benchmark problems consists of $2,760$ data stream benchmarks. 

    \item \textbf{Drift locality impact evaluation.} We present a comprehensive study designed to assess the influence of concept drift locality on its detection, encompassing both binary and multi-class data streams.
    
    \item \textbf{Comparison between state-of-the-art drift detectors.} We conduct an extensive, comprehensive, and reproducible comparative study among $9$ state-of-the-art drift detectors based on the proposed framework and $2,760$ benchmark problems.
    
    \item \textbf{Recommendations and open challenges.} Based on the results from the experimental study, we derive recommendations seeking insights into the strengths and weaknesses of the top-performing drift detectors. These recommendations aim to provide a comprehensive understanding of the detectors' capabilities. Additionally, we identify open challenges within the domain of learning from data streams impacted by concept drift, outlining directions for future research.
\end{itemize}



This paper is structured as follows: Section~\ref{sec:background} provides the theoretical foundation and discusses related work. Section~\ref{sec:taxonomy} introduces our proposed concept drift locality categorization and presents the benchmark problems. Section~\ref{sec:exp_setup} outlines the experimental setup, while Section~\ref{sec:results} presents the results. Section~\ref{sec:lessons} delves into the lessons learned, and Section~\ref{sec:conclusions} offers the conclusion and outlines directions for future work.

\section{Background and related work}
\label{sec:background}
This section reviews the background and related work. We provide an overview of the literature on data streams and concept drift detection.

\subsection{Data streams}

A data stream refers to a potentially unbounded sequence of ordered instances that arrive over time within a system. Learning from data streams imposes specific limitations on classifiers~\cite{gama2010knowledge}. We can define a stream, denoted as S, as a sequence $<s_1, s_2, s_3, . . . , s_{\infty} >$, where $s_i = (X, y)$. This stream can be handled either one instance at a time (online scenario) or in batches (block scenario). Data streams exhibits four primary characteristics~\cite{bahri2021data, aguiar2022survey}: (i) Volume, (ii) Velocity, (iii) Veracity, and (iv) Non-stationarity, which present challenges to classifiers that must adapt accordingly. 

\begin{figure}[t!]
    \centering
    \includegraphics[width=\textwidth]{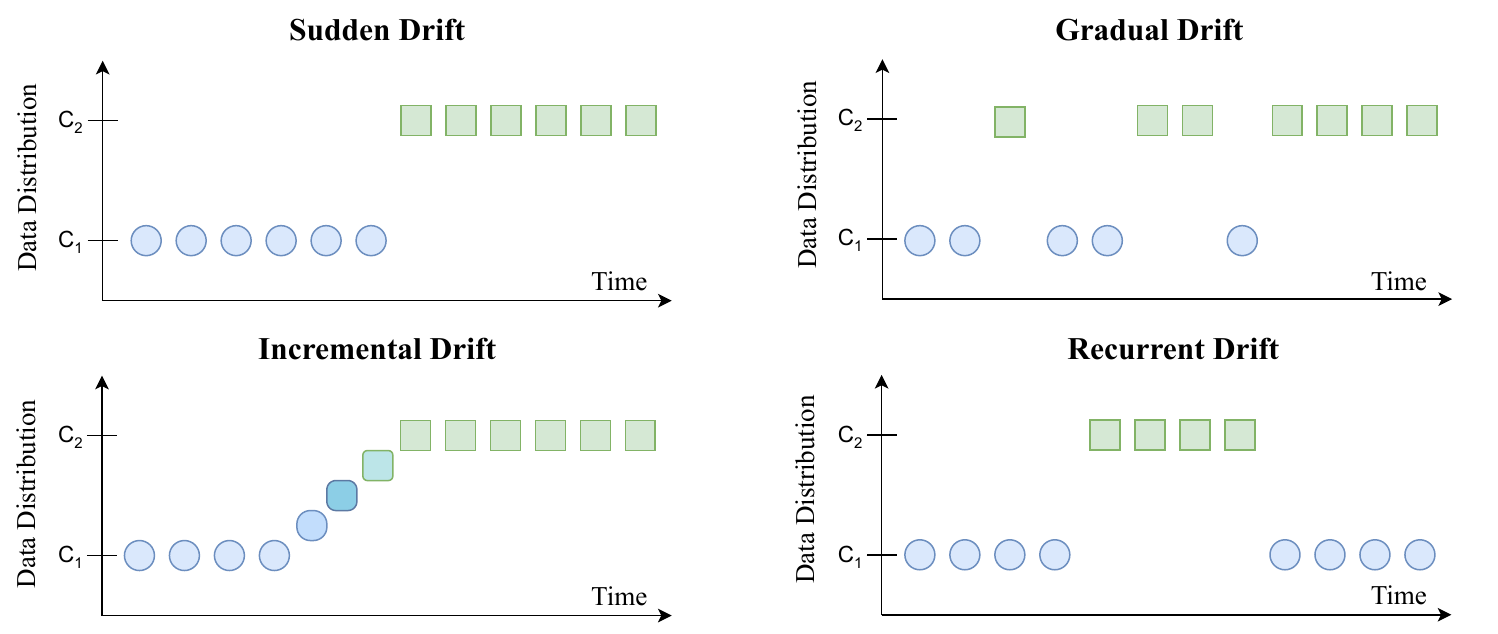}
    \caption{Different types of drifts depending on their speed.}
    \label{fig:concept_drift_example}
\end{figure}

Data streams are susceptible to a phenomenon known as concept drift~\cite{krawczyk2017ensemble, lu2018learning}. Each instance arrives at a specific time, denoted as t, and is generated based on a probabilistic distribution denoted as $\phi^t (X, y)$, where $X$ represents the feature vector and $y$ denotes the class label. If all instances in the stream are generated by the same probability distribution, the data is considered stationary, indicating a single and stable underlying concept. However, in real-world applications, data rarely adheres to stationary assumptions~\cite{masegosa2020analyzing}. On the other hand, when two distinct instances arriving at times $t$ and $t+C$ are generated by $\phi^t (X, y)$ and $\phi^{t+C} (X, y)$ respectively, and if $\phi^t \neq \phi^{t+C}$, indicates the occurrence of a concept drift. This phenomenon impacts varios aspects of a data stream and, as such can be examined from multiple viewpoints. When analyzing and comprehending concept drift, the following factors come into consideration~\cite{korycki2021concept, aguiar2022survey}:

\begin{itemize}
\item \textbf{Influence of the decision boundaries.} Firstly, it is necessary to consider how concept drift affects the learned decision boundaries, distinguishing between real and virtual concept drifts. Virtual drift produces a change in the unconditional probability distribution $P(x)$, without affecting the learned decision boundaries. Although virtual drift does not impair learning models, its detection is necessary to avoid false alarms and prevent unnecessary, costly adaptations. In contrast, real concept drift modifies the decision boundaries, making them worthless to the current concept. Detecting and adapting to real concept drift is crucial for preserving predictive performance. 

\item \textbf{Speed of changes.} There are three types of concept drift~\cite{webb2016characterizing}: (i) incremental; (ii) gradual; and (iii) sudden concept drifts as illustrated in Fig.~\ref{fig:concept_drift_example}. Incremental drift generates a sequence of intermediate states between the old and new concept, while gradual drift oscillates between instances coming from both old and new concepts, with the new concept becoming more and more frequent over time. Finally, sudden drift instantaneously switches between old and new concepts, leading to an instant degradation of the underlying learning algorithm.

\item \textbf{Recurrence.} Changes in the stream can be either unique or recurring. In the latter case, the previously seen concept may reemerge over time, allowing us to recycle previously learned knowledge.  The past knowledge can be used as an initialization point for the drift recovery.

\item \textbf{Presence of noise.} Noise can take the form of sporadic, insignificant variations within a stream that can be disregarded, or substantial corruption within the features or class labels that need to be dealt with in order to prevent the input of misleading or adversarial data into the classifier~\cite{krawczyk2018online}. Drift detectors must ignore noise.

\item \textbf{Locality}. The literature distinguishes between global and local concept drifts~\cite{gama2006learning}. The former impacts the entire data stream, while the latter pertains to specific parts of it, such as individual clusters of instances or subsets of classes. However, this classification is too shallow and a more in-depth and detailed discussion regarding the concept drift's locality will be introduced in this manuscript.

\end{itemize}


To address the challenges posed by concept drift, two approaches are commonly employed: (i) implicit and (ii) explicit. Implicit methods manage drift adaptation through intrinsic mechanisms integrated within the classifier, assuming its capability to self-adjust to new instances reflecting the most recent concept while gradually discarding outdated information~\cite{ditzler2015learning, suarez2022survey}. These approaches involve establishing appropriate learning and forgetting rates, utilizing adaptive sliding windows, or continually tuning hyperparameters. Conversely, explicit approaches assign drift adaptation to an external tool known as a drift detector~\cite{suarez2022survey, lu2018learning}. Drift detectors continuously monitor stream properties (e.g., statistics) or classifier performance (e.g., error rates). They raise a warning signal when there are indications of impending drift and trigger an alarm signal when concept drift has occurred.

\subsection{Drift detectors}
In recent years, a plethora of explicit drift detectors have been introduced~\cite{barros2018large}. As defined by \citet{lu2018learning}, drift detectors function by extracting critical features from both historical and newly arrived data and subsequently subjecting them to dissimilarity tests. Like classifiers, drift detection methods fall into three categories: supervised (or error rate-based), semi-supervised, and unsupervised (or data distribution-based)\cite{lu2018learning}. The primary distinction among these categories is the point at which drift is identified. While supervised drift detectors pinpoint alterations in class boundaries, unsupervised detectors focus on tracking shifts in data distribution~\cite{gama2014survey}.

Among supervised drift detectors, the most widely adopted group relies on evaluating classifier error or accuracy using labeled instances. This category of drift detectors can be further divided into three distinct types: (i) change detection-based detectors, (ii) statistical-based detectors, and (iii) window-based detectors.
\begin{itemize}

\item \noindent \textbf{Change detection-based detectors}. In the first category, we find detectors like Page Hinkley~\cite{page1954continuous}, CUSUM~\cite{page1954continuous}, and Geometric Moving Average~\cite{roberts2000control}. These detectors utilize cumulative sums to trigger an alarm when a significant change in input data occurs. While they are computationally efficient, their performance heavily depends on the selection of hyperparameters.

\item \noindent \textbf{Statistical-based detectors.} The second group comprises detectors such as the Drift Detector Method (DDM)~\cite{gama2004learning} and its variants like Early DDM (EDDM)~\cite{baena2006early}, Reactive DDM (RDDM)~\cite{barros2017rddm}, DDM based on Hoeffding's bound (HDDM)~\cite{frias2014online}, Wilcoxon Rank Sum Test Drift Detector (WSTD)~\cite{de2018wilcoxon}, and EWMA for Drift Detection (ECDD)~\cite{ross2012exponentially}. These methods compute statistical features over time based on error rates. They operate under the assumption that as long as the data distribution remains stationary, the learner's error rate will decrease with an increasing number of analyzed samples. When the error rate exceeds a predefined threshold, an alert is triggered.

\item  \noindent \textbf{Window-based detectors.} Finally, a popular category of supervised detectors relies on metrics calculated within subwindows of a data stream. An example is the ADaptative WINdow (ADWIN)~\cite{bifet2007learning}, which employs an adaptive sliding window based on Hoeffding's inequality. ADWIN manages a sliding window divided into two sub-windows representing old and new data and dynamically adjusts its size, expanding during periods of stability and contracting in the presence of drift. ADWIN signals a drift when the average between the two windows surpasses a predefined threshold. This approach has inspired the development of several other detectors such as Kolmogorov-Smirnov Windowing (KSWIN)~\cite{raab2020reactive}, Statistical Test of Equal Proportions (STEPD)~\cite{nishida2007detecting}, and others~\cite{huang2014detecting, pesaranghader2016fast}. 
    
\end{itemize}


In addition to detectors based on classifier accuracy, there are also supervised trainable detectors that employ traditional machine learning classifiers to detect concept drift. Examples of such detectors include the Restricted Boltzmann Machine (RBM-IM)~\cite{korycki2021concept}, the Complexity Drift Detector (C2D)~\cite{komorniczak2023complexity} and QuadCDD~\cite{wang2024quadcdd}. It is important to note that there is a notable absence of drift detectors specifically designed for multi-class scenarios, which come with their own unique characteristics and challenges.

Unsupervised drift detectors are primarily focused on identifying disparities in unlabeled data without the need for added supervision. These detectors typically employ a distance metric to quantify the dissimilarity between the distribution of historical data and newly arrived data~\cite{lapinski2018empirical, sobolewski2013comparable}. When this dissimilarity is statistically significant, it triggers a process to update the learning model.
In this category, we find detectors such as Statistical Change Detection for multi-dimensional data (SCD)\cite{song2007statistical}, the PCA-based change detection framework (PCA-CD)\cite{qahtan2015pca}, Equal Density Estimation (EDE)\cite{gu2016concept}, Least Squares Density Difference-based Change Detection Test (LSDD-CDT)\cite{bu2016pdf}, among others~\cite{lu2018learning}. More advanced unsupervised methods aim to precisely locate where the drift occurred in the feature space. These detectors focus on spatial searches employing various dissimilarity measures~\cite{liu2017regional, liu2018accumulating}. They tackle concept drift at its root, addressing distribution drift. Typically, these algorithms require users to define both a historical time window and a new data window. A common approach is to use two sliding windows, with the historical time window fixed while sliding the new data window~\cite{lu2014concept}.

As aforementioned, numerous drift detectors have been proposed in the literature, highlighting the need for comparisons to understand how each one behaves in different complex scenarios. While \citet{lu2018learning} and \citet{suarez2022survey} reviewed the state-of-the-art in data stream learning with a focus on drift detectors, however they lacked an empirical comparison between these detectors.  Additionally, the locality of the concept drift is often neglected~\cite{korycki2021concept, gama2006learning}. In a different approach, \citet{barros2018large} conducted an extensive empirical comparison and evaluation of concept drift detection methods across various data stream configurations. Moreover, Table~\ref{tab:related_works} presents a summary of works that empirically compare drift detectors and the type of concept drift evaluated. However, a notable gap exists in the literature concerning concept drift detection in data streams with multiple classes and how the locality of the drift influences its detection.

\begin{table}[]
\centering
\caption{Comparison of most commonly applied, drift detectors and types of concept drift by
contributions in related works}
\label{tab:related_works}
\resizebox{.6\textwidth}{!}{%
\begin{tabular}{@{}lll@{}}
\toprule
\textbf{Reference} & \textbf{Drift Detectors} & \textbf{Drift Type} \\ \midrule
\citet{gama2004learning} & DDM & Sudden; Gradual \\ \midrule

\citet{baena2006early} & DDM, EDDM  & Sudden; Gradual \\ \midrule

\citet{frias2014online} & DDM, ADWIN, HDDM$_{W}$, ECDD & Sudden; Gradual \\ \midrule

\citet{gonccalves2014comparative} & \begin{tabular}[c]{@{}l@{}} DDM, EDDM, ADWIN, ECDD,\\ STEPD, PH, NB, PL, DOF \end{tabular} & Sudden; Gradual \\ \midrule

\citet{barros2018large} & \begin{tabular}[c]{@{}l@{}}DDM, EDDM, ADWIN, HDDM$_W$, \\ HDDM$_A$,  ECDD, SEQDRIFT, SEED, \\ STEPD, FHDDM, FTDD, RDDM, WSTD\end{tabular} & Sudden; Gradual \\ \midrule

\citet{santos2019differential} & \begin{tabular}[c]{@{}l@{}}DDM, EDDM, ADWIN, HDDM$_W$, \\ HDDM$_A$, ECDD, SEQDRIFT, STEPD, \\ FHDDM, RDDM, and WSTD\end{tabular} & Sudden; Gradual \\ \midrule

\citet{baburouglu2021novel} & \begin{tabular}[c]{@{}l@{}} DDM, EDDM, ADWIN, HDMM$_W$,\\ HDDM$_A$, ECDD, SEQDRIFT, SEED, \\ STEPD, FHDDM, FTDD, RDDM, \\ WSTD, GMA, PH\end{tabular} & Sudden \\ \midrule

\citet{korycki2021concept} & \begin{tabular}[c]{@{}l@{}}FHDDM, RDDM, WSTD, PerfSim,\\ DDM-OCI, RBM-IM \end{tabular} & \begin{tabular}[c]{@{}l@{}}Sudden; Gradual; \\ Incremental \end{tabular} \\ \midrule

\citet{poenaru2022concept} & \begin{tabular}[c]{@{}l@{}}DDM, EDDM, ADWIN, HDDM$_W$,\\ HDDM$_A$\end{tabular} & Sudden; Gradual \\ \midrule

\citet{sakurai2023benchmarking} &  DDM, EDDM, HDDM$_W$, HDDM$_A$ & \begin{tabular}[c]{@{}l@{}}Sudden; Gradual; \\ Incremental \end{tabular} \\ \bottomrule
\end{tabular}%
}
\end{table}

\section{Proposed categorization of concept drift locality}
\label{sec:taxonomy}

As discussed in various studies~\cite{korycki2021concept,gama2006learning, brzezinski2021impact, lango2022makes}, local data difficulty factors and the number of classes exert a substantial influence on the capabilities of classifiers when handling data streams. Furthermore, these factors also play a major role in concept drift detection. Notably, the performance of standard drift detectors can vary significantly based on the underlying distribution of the stream, especially when they lack strategies to address these factors. This can result in either excellent performance or suboptimal outcomes.

However, the current body of literature addressing the categorization and characterization of concept drifts in streaming classification problems often overlooks two crucial factors: the number of classes affected and local data difficulties. It is important to note that a significant portion of concept drift research primarily focuses on binary classification and global shifts in data distribution, or perturbations that primarily impact the minority class, as observed in the imbalanced scenario~\cite{brzezinski2021impact}. Given the substantial impact that local data difficulty factors and the number of classes affected can have on non-stationary data streams, the existing categorizations of concept drift may fall short in evaluating concept drift detection in dynamic environments. They may fail to encompass all the nuances that make a concept drift particularly challenging or solvable. Therefore, we propose an extended concept drift categorization that explicitly incorporates considerations of locality and the number of classes affected when assessing concept drift. This expanded framework should pave the way for further research in the development of classifiers and drift detectors tailored for multi-class data streams in non-stationary environments, enabling systematic studies under different relevant drifting conditions.

\subsection{Categorization of concept drift locality}
First, we formally define the proposed categorization. Consider a bounded $d$-dimensional attribute space $X$ and output space $y$ and a given posterior probability distribution $\phi^t (X, y)$ at time $t$, which can be expanded as $\phi_{c_1}^t (X, y) \cup \phi_{c_2}^t (X, y) \dots \cup \phi_{c_n}^t (X, y)$ where $c_i$ represents the distribution for a given class and $n$ the number of classes. As previously defined if  $\phi^t (X, y) \neq \phi^{t+C} (X, y)$ a concept drift has occurred and falls into one of the four categories:
\begin{itemize}
    \item \textbf{Single-Class Local Concept Drift}: Given the drifted distribution $\phi^{t+C} (X, y)$ exists \textbf{only one} $i \in [0,n]$ that $\phi_{c_i}^{t+C} (X, y) = \phi_{c_i}^{t} (X, y) \cup \hat{\phi_{c_i}} (X,y)$.
    \item \textbf{Single-Class Global Concept Drift}: Given the drifted distribution $\phi^{t+C} (X, y)$ exists \textbf{only one} $i \in [0,n]$ that $\phi_{c_i}^{t+C} (X, y) = \hat{\phi_{c_i}} (X,y)$.
    \item \textbf{Multi-Class Local Concept Drift}: Given the drifted distribution $\phi^{t+C} (X, y)$ exists \textbf{more than one} $i \in [0,n]$ that $\phi_{c_i}^{t+C} (X, y) = \phi_{c_i}^{t} (X, y) \cup \hat{\phi_{c_i}} (X,y)$
    \item \textbf{Multi-Class Global Concept Drift}: Given the drifted distribution $\phi^{t+C} (X, y)$ exists \textbf{more than one} $i \in [0,n]$ that $\phi_{c_i}^{t+C} (X, y) = \hat{\phi_{c_i}} (X,y)$.
\end{itemize}

\begin{figure}[t!]
    \centering
    \begin{subfigure}[b]{.7\columnwidth}
        \includegraphics[width=\columnwidth]{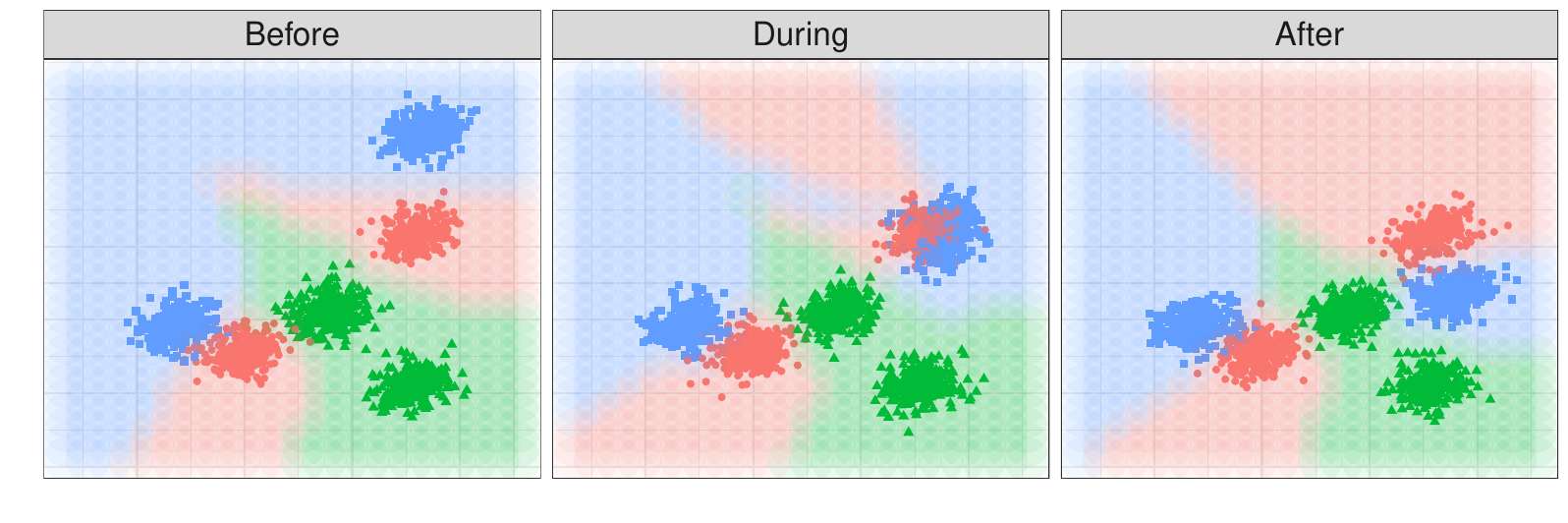}
        \caption{Single-Class Local}
    \end{subfigure}
    \begin{subfigure}[b]{.7\columnwidth}
        \includegraphics[width=\columnwidth]{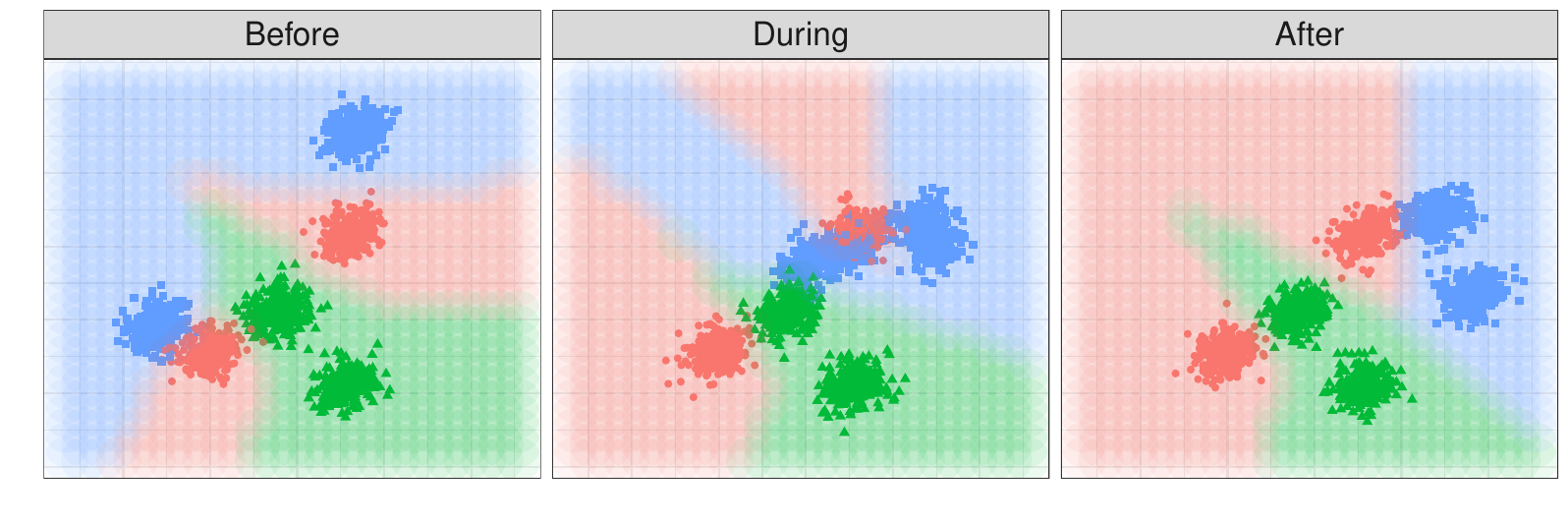} 
        \caption{Single-Class Global}
    \end{subfigure}
    
    \caption{Illustrative data distribution before, during and after a Single-Class Local (a) and Global (b) concept drift. Each color represents one class and background color represents decision boundaries.}
    \label{fig:categorization_single}
\end{figure}

\begin{figure}[t!]
    \centering
    \begin{subfigure}[b]{.7\columnwidth}
        \includegraphics[width=\columnwidth]{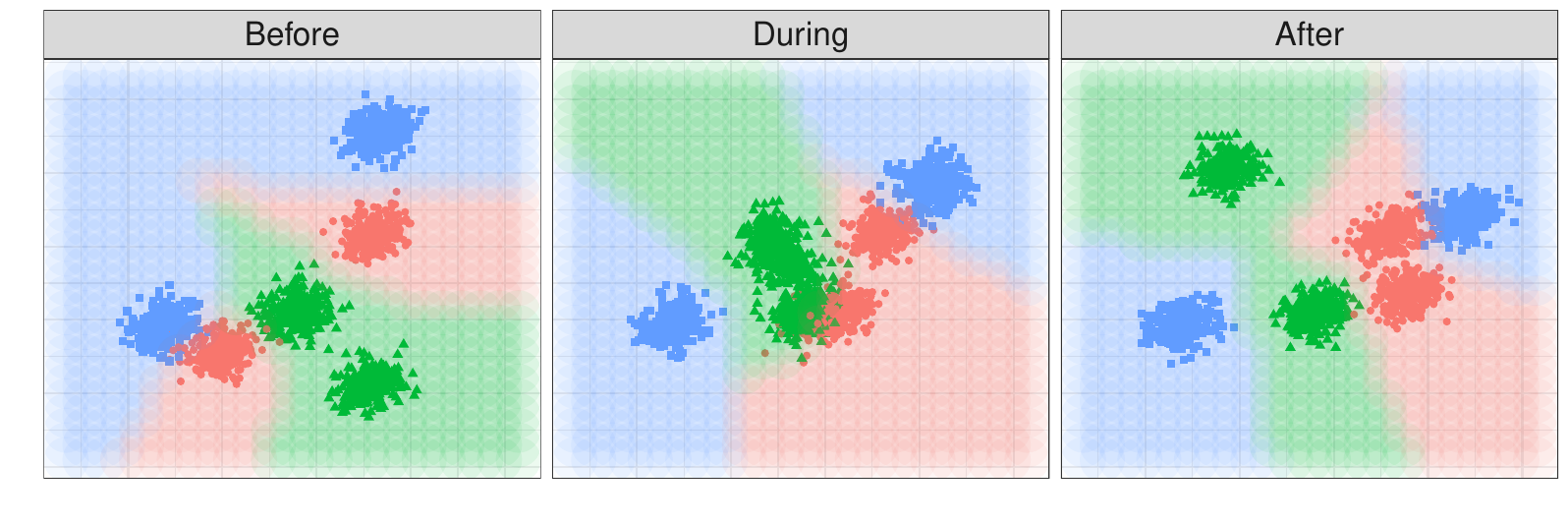}
        \caption{Multi-Class Local}
    \end{subfigure}
    \begin{subfigure}[b]{.7\columnwidth}
        \includegraphics[width=\columnwidth]{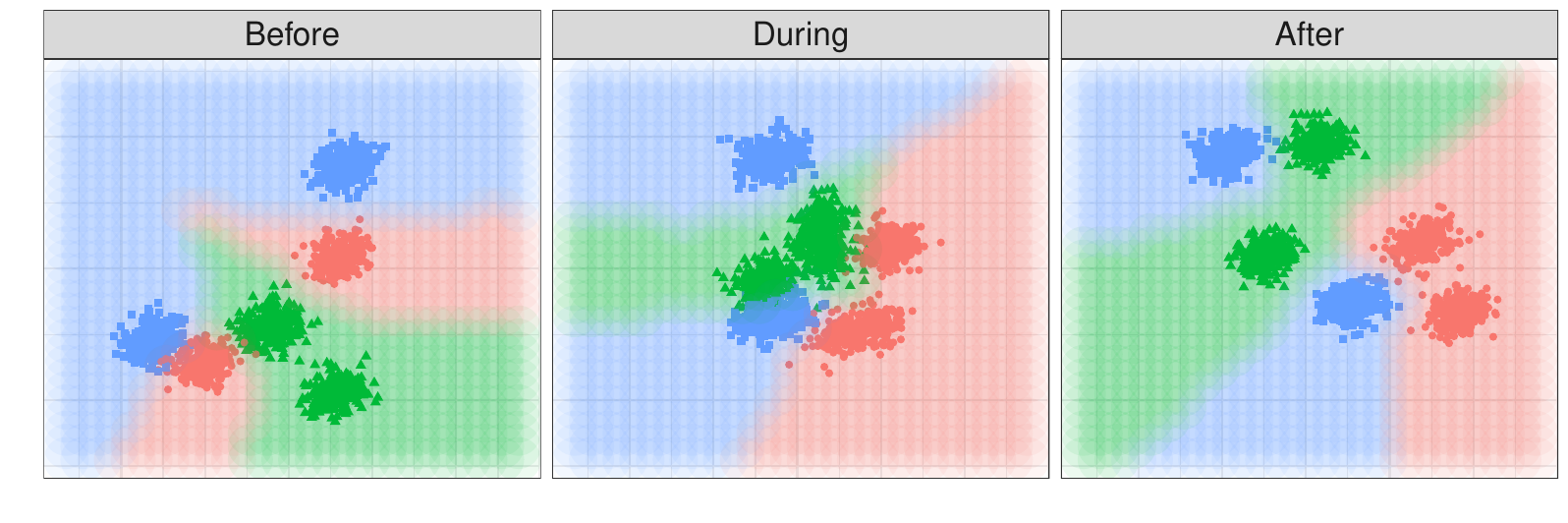}
        \caption{Multi-Class Global}
    \end{subfigure}
     \caption{Illustrative data distribution before, during and after a Multi-Class Local (a) and Global (b) concept drift. Each color represents one class and background color represents decision boundaries.}
    \label{fig:categorization_multi}
\end{figure}

Figs.~\ref{fig:categorization_single} and \ref{fig:categorization_multi} present a visual representation of the four defined categories. Local drifts occur when the original concept and a new distribution emerge without overlapping. This type of concept drift can be exceedingly subtle yet has the potential to compromise predictive accuracy in specific regions of the feature space. Consequently, it poses a challenge for detection and resolution. On the other hand, global concept drift involves a complete transformation of the data distribution, often resulting in less subtle changes. However, the ease of drift detection depends on the behavior of the new distribution.
It is worth noting that learning mechanisms and concept drift detection strategies can differ significantly in scenarios with multiple classes~\cite{korycki2021concept, lango2022makes}. Thus, when assessing classifier performance or concept drift detection, we must also account for the number of classes affected by the concept drift. This proposed categorization enables us to analyze concept drift detection while considering intricacies that have not received extensive attention in the existing literature. Additionally, it empowers  the development of novel classifiers and drift detectors capable of handling a spectrum of concept drifts, ranging from subtle to noticeable.

\subsection{Benchmarks}
To explicitly assess the performance of classifiers and drift detectors in data streams featuring the concept drift categories outlined earlier, we introduce a set of drift difficulties corresponding to each category within our proposed framework. These difficulties were implemented using two widely recognized data stream generators, namely Random RBF and Random Tree, as established in the literature~\cite{aguiar2022survey}.
Table~\ref{tab:spec_benchmark} presents a comprehensive overview of the proposed data difficulties along with their respective categorizations. For each difficulty setting, we generated data streams encompassing a range of class counts, including $\{2, 3, 5, 10\}$ classes, and varying feature dimensions of $\{2, 5, 10\}$. Additionally, we introduced three distinct types of drifts: Sudden, Gradual, and Incremental.
In the context of Multi-Class drifts, we considered scenarios where drifts affected subsets of classes, including $\{\{2\}, \{2, 3\}, \{2, 3, 5\}, \{2, 3, 5, 10\}\}$, depending on the number of classes. 

Finally, to ensure reproducibility and facilitate the utilization of our proposed categorization and benchmarks, all the data streams used in our experiments have been made publicly available for future research.~\footnote{\href{https://github.com/gabrieljaguiar/locality-concept-drift}{https://github.com/gabrieljaguiar/locality-concept-drift}}

\begin{sidewaystable*}
    \caption{Concept drift locality benchmark specifications. S: Sudden, G: Gradual, I: Incremental.}
    \label{tab:spec_benchmark}
    \resizebox{\textwidth}{!}{%
    \begin{tabular}{@{}ccclll@{}}
    \toprule
    \textbf{Generator} & \textbf{\begin{tabular}[c]{@{}c@{}}Impact on\\  the class space\end{tabular}} & \textbf{\begin{tabular}[c]{@{}c@{}}Impact on \\ the feature space\end{tabular}} & \textbf{Difficulty} & \textbf{Description} & \textbf{Drift Speed} \\ \midrule
    \multirow{21}{*}{RBF} & \multirow{10}{*}{\begin{tabular}[c]{@{}c@{}}Single-\\ Class\end{tabular}} & \multirow{5}{*}{Local} & emerging\_cluster & A new subcluster emerge & S/G \\
     &  &  & reappearing\_cluster & Part of the clusters of ONE class disappears for a period of time & S/G \\
     &  &  & splitting\_cluster & Part of the clusters of ONE class splits into two clusters that move to another direction & S/G/I \\
     &  &  & merging\_cluster & Part of the clusters of ONE class clusters are merged in the midpoint between them & S/G/I \\
     &  &  & moving\_cluster & Part of the clusters of ONE class center moves to another position & S/G/I \\
     &  & \multirow{5}{*}{Global} & reappearing\_cluster & All clusters of ONE class disappear for a period of time & S/G \\
     &  &  & splitting\_subcluster & All subclusters of ONE class splits into two subclusters that move to another direction & S/G/I \\
     &  &  & merging\_cluster & All subclusters of ONE class of one class merge & S/G/I \\
     &  &  & moving\_cluster & All clusters of ONE class moves to another position & S/G/I \\
     &  &  & class\_emerging & A new class appear & S/G \\
     & \multirow{11}{*}{\begin{tabular}[c]{@{}c@{}}Multi-\\ Class\end{tabular}} & \multirow{6}{*}{Local} & emerging\_cluster & A new emerging subcluster for N classes appears & S/G \\
     &  &  & reappearing\_cluster & Part of the clusters of N classes disappears for a period of time & S/G \\
     &  &  & splitting\_cluster & Part of the clusters of N classes splits into two clusters that move to another direction & S/G/I \\
     &  &  & merging\_clusters & Part of the clusters  clusters are merged in the midpoint between them for N classes & S/G/I \\
     &  &  & moving\_cluster & Part of the clusters of N classes center moves to another position & S/G/I \\
     &  &  & swap\_cluster & Part of the clusters of N classes swap position & S/G \\
     &  & \multirow{5}{*}{Global} & reappearing\_cluster & All clusters of N classes disappear for a period of time & S/G \\
     &  &  & splitting\_cluster & All clusters of N classes splits into two clusters that move to another direction & S/G/I \\
     &  &  & merging\_cluster & All clusters of N classes center merges & S/G/I \\
     &  &  & moving\_cluster & All clusters of N classes moves to another position & S/G/I \\
     &  &  & swap\_cluster & All clusters of N classes swaps position & S/G \\ \midrule
    \multirow{15}{*}{RT} & \multirow{6}{*}{\begin{tabular}[c]{@{}c@{}}Single-\\ Class\end{tabular}} & \multirow{3}{*}{Local} & emerging\_branch & A new branch appears & S/G \\
     &  &  & prune\_regrowth\_branch & Part of the branches of ONE class are pruned and then regrowth this branches & S/G \\
     &  &  & prune\_growth\_new\_branch & Part of the branches of ONE class are pruned and then other branches grow & S/G \\
     &  & \multirow{3}{*}{Global} & prune\_regrowth\_branch & All of the branches of ONE class are pruned and then regrowth this branches & S/G \\
     &  &  & prune\_growth\_new\_branch & All of the branches of ONE class are pruned and then other branches grow & S/G \\
     &  &  & class\_emerging & A new class appear & S/G \\
     & \multirow{9}{*}{\begin{tabular}[c]{@{}c@{}}Multi-\\ Class\end{tabular}} & \multirow{5}{*}{Local} & emerging\_branch & A new branch appears for N classes & S/G \\
     &  &  & prune\_regrowth\_branch & Part of the branches of N classes are pruned and then regrowth this branches & S/G \\
     &  &  & prune\_growth\_new\_branch & Part of the branches of N classes are pruned and then other branches grow & S/G \\
     &  &  & split\_node & Part of the leaves of N classes are splited to generate two leafs of two different classes & S/G \\
     &  &  & swap\_leaves & Part of the leaves of N classes swaps with another class & S/G \\
     &  & \multirow{4}{*}{Global} & prune\_regrowth\_branch & All of the branches of N classes are pruned and then regrowth this branches & S/G \\
     &  &  & prune\_growth\_new\_branch & All of the branches of N classes are pruned and then other branches grow & S/G \\
     &  &  & split\_node & All of the leaves of N classes are splited to generate two leafs of two different classes & S/G \\
     &  &  & swap\_leaves & All of the leaves of N classes swaps with another class & S/G \\ \bottomrule
    \end{tabular}%
    }

\end{sidewaystable*}

\begin{figure}[b!]
    \centering
    \begin{subfigure}[b]{.7\columnwidth}
        \includegraphics[width=\columnwidth]{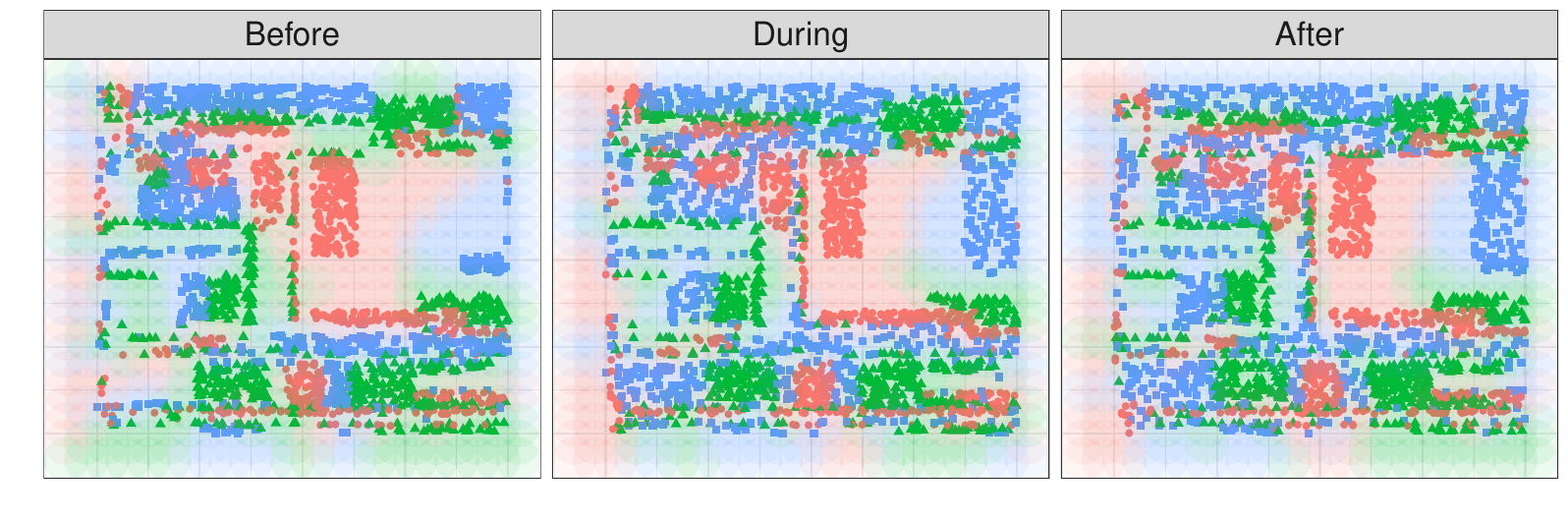}
        \caption{Single-Class Local}
    \end{subfigure}
    \begin{subfigure}[b]{.7\columnwidth}
        \includegraphics[width=\columnwidth]{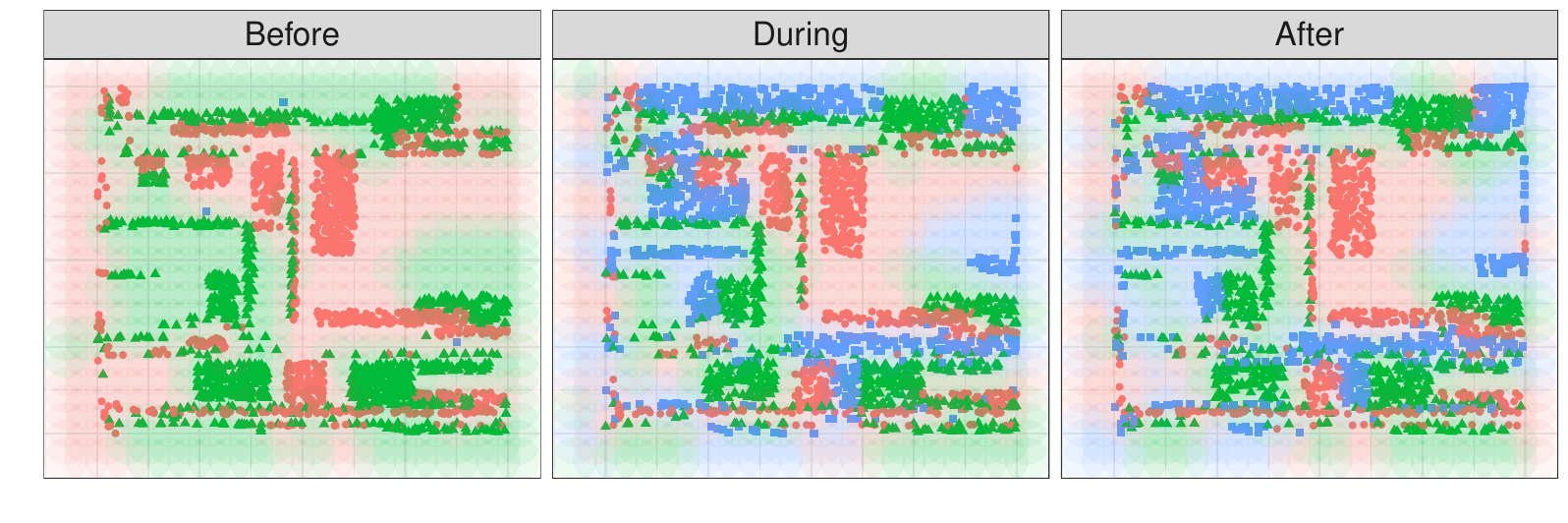}
        \caption{Single-Class Global}
    \end{subfigure}

    \caption{Data distribution before, during and after a Single-Class Local (a) and Global (b) \texttt{emerging\_branch} concept drift. Each color represents one class and background color represents decision boundaries.}
    \label{fig:single_class_examples}
\end{figure}

\begin{figure}[t!]
    \centering
    \begin{subfigure}[b]{.75\columnwidth}
        \includegraphics[width=\columnwidth]{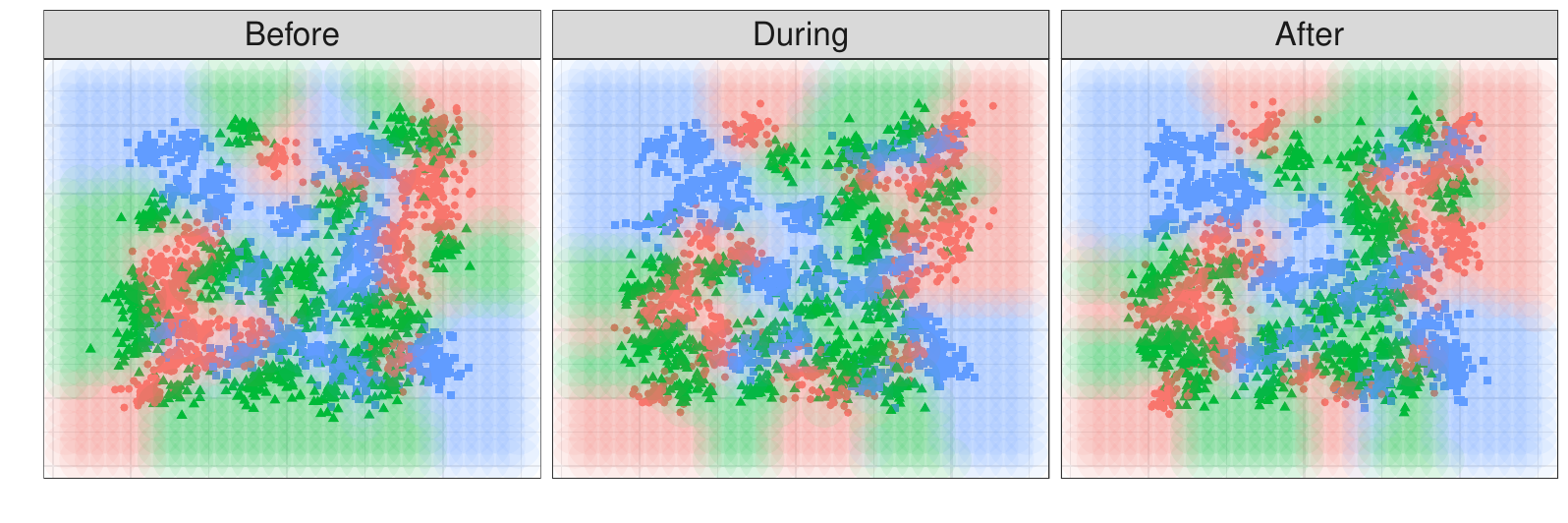}
        \caption{Multi-Class Local}
    \end{subfigure}
    \begin{subfigure}[b]{.75\columnwidth}
        \includegraphics[width=\columnwidth]{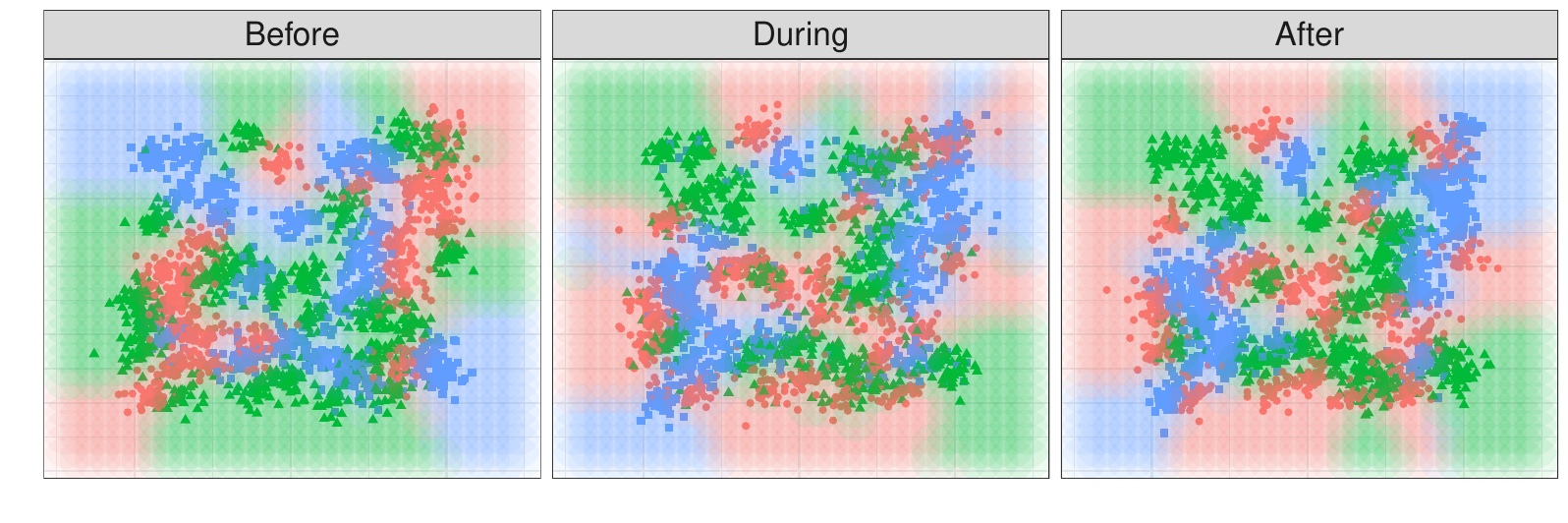}
        \caption{Multi-Class Global}
    \end{subfigure}
    \caption{Data distribution before, during and after a Multi-Class Local (a) and Global (b) \texttt{swap\_cluster} concept drift with all classes affected. Each color represents one class and background color represents decision boundaries.}
    \label{fig:multi_class_examples}
\end{figure}

\clearpage
\section{Experimental Setup}
\label{sec:exp_setup}

The experiments are designed to assess the performance of drift detectors across a diverse range of drift challenges. We also investigate how the locality and the number of affected classes influence drift detection. The primary goal is to gain insights into the performance of various drift detectors and how each type of drift challenge affects their effectiveness. Additionally, we examine the impact of each challenge on classifiers that rely on explicit drift detectors. To address these objectives, we formulate the following research questions (RQ):

\begin{itemize}
    \item \textbf{RQ1}: Which concept drift detector demonstrates the most effective detection performance across all scenarios?
    \item \textbf{RQ2}: Which concept drift detector excels in detection performance for each distinct type of scenario?
    \item \textbf{RQ3}: How does each category of concept drift influence the performance of the drift detector?
    \item \textbf{RQ4}: Which specific difficulty presents the most challenging scenario?
    \item \textbf{RQ5}: How does the number of classes and features affect the concept drift detection?
    \item \textbf{RQ6}: How does each scenario impact the performance of classifiers utilizing the best-performing drift detector?
\end{itemize}


\subsection{Drift detectors}

To assess the performance of drift detectors across various types of concept drift, we chose $9$ state-of-the-art supervised drift detectors displayed in Table~\ref{tab:drift_detectors}. We made these selections based on their widespread use in the literature and their demonstrated effectiveness in numerous scenarios. It is important to note that we included at least one representative from each group of supervised drift detectors mentioned in Section~\ref{sec:background}. This choice was deliberate, aiming to gain insights into the strengths and weaknesses of each group.

\begin{table}[b!]
\centering
\caption{Drift detectors used in the experiments.}
\label{tab:drift_detectors}
\resizebox{.6\textwidth}{!}{%
\begin{tabular}{@{}lll@{}}
\toprule
\textbf{Acronym} & \textbf{Name} & \textbf{Reference} \\ \midrule
ADWIN & ADaptive Windowing & \citet{bifet2007learning} \\
KSWIN & Kolmogorov-Smirnov Windowing & \citet{raab2020reactive} \\
STEPD & Statistical Test of Equal Proportions & \citet{nishida2007detecting} \\
DDM & Drift Detection Method & \citet{gama2004learning} \\
EDDM & Early Drift Detection Method & \citet{baena2006early} \\
HDDM & Drift Detection Method based on Hoefing bound & \citet{frias2014online} \\
RDDM & Reactive Drift Detection Method & \citet{barros2017rddm} \\
ECDD & EWMA for Drift Detection &  \citet{ross2012exponentially} \\
PH & Page Hinkley & \citet{page1954continuous} \\ \bottomrule
\end{tabular}%
}
\end{table}

To monitor drift, we leverage the error distribution of the classifier to effectively monitor concept drifts, as suggested in the existing concept drift literature~\cite{bifet2007learning}. The error distribution is binary, and we determine the value to update the drift detector using the equation:

\begin{equation}
    \label{eq:error_distribution}
    \phi =
    \begin{cases}
        1, \text{ if } L(x) = y \\
        0, \text{ if } L(x) \neq y \\
    \end{cases}
\end{equation}

\noindent where $\phi$ represents the error distribution, $x$ represents a new instance, $y$ denotes the correct class, and $L$ the classifier. By employing this approach, we can effectively detect concept drifts when there are changes in the classifier error distribution.

\subsection{Classifier}

To evaluate the performance of the drift detectors, we opted to use Hoeffding Tree (HT)~\cite{holmes2005stress} as our classifier. This classifier is known for its exceptional predictive performance and lacks inherent drift detection or adaptation mechanisms, ensuring an unbiased assessment of concept drift detection.

\subsection{Performance evaluation}

Given that each data stream in our experiments is characterized by a single known drift point, we employed an evaluation process similar to that used in classification tasks. Consequently, we extracted True Positive (TP), False Positive (FP), and False Negative (FN) values according to the following definitions:

\begin{itemize}
    \item \textbf{TP}: When a concept drift detected at time $t$ and $t \in [i_d, i_d + R]$ where $i_d$ is the instance where the concept drift happened, and $R$ is the range of detection.
    \item \textbf{TN}: When no alert is raised and there is no known concept drift.
    \item \textbf{FP}: When a concept drift alert is raised at $t$ and $t \not\in [i_d, i_d + R]$ where $i_d$ is the instance where the concept drift happened, and $R$ is the range of detection..
    \item \textbf{FN}: When no alert is raised within the range of detection $R$.
\end{itemize}

Subsequently, we calculated three performance metrics: Precision (Eq.~\ref{eq:precision}), Recall (Eq.~\ref{eq:recall}), and F1-Score (Eq.~\ref{eq:f1}). Precision assesses the ratio between false alarms and accurate drift alerts, while recall measures the ratio between correct drift detection and undetected drifts. The F1-Score combines both of these metrics. These three metrics in combination provide insights into the behavior of each concept drift detector.

\hspace{-0.9cm}
\begin{minipage}[h!]{0.5\columnwidth}
\begin{equation} 
    \label{eq:precision}
    Precison = \displaystyle\frac{TP}{TP + FP}
\end{equation}
\end{minipage}
\begin{minipage}[h!]{0.5\columnwidth}
\begin{equation}
    \label{eq:recall}
    Recall = \frac{TP}{TP + FN}
\end{equation}
\end{minipage}
\begin{equation}
    \label{eq:f1}
    F1-Score = \frac{2*(Precision*Recall)}{Precision + Recall}
\end{equation}

Furthermore, for each accurate drift alarm, we calculate the average detection delay as described in Eq.~\ref{eq:delay}, where $D$ represents the instance position of the drift, and $D_d$ indicates the instance position when the drift alarm was triggered. This metric assesses the speed of drift detection.

\begin{equation}
    \label{eq:delay}
    Delay = \frac{\sum (D_d - D)}{TP}
\end{equation}

\subsection{Reproducibility}

The source code of the algorithms, the experiments and the data streams are publicly available on GitHub to facilitate the transparency and reproducibility of this research~\footnote{\href{https://github.com/gabrieljaguiar/locality-concept-drift}{https://github.com/gabrieljaguiar/locality-concept-drift}}. All results, interactive plots and tables are available
on the website~\footnote{Interactive plots and tables for all experiments are available at \href{https://gabrieljaguiar.github.io/comprehensive-concept-drift/}{https://gabrieljaguiar.github.io/comprehensive-concept-drift/}}. All the experiments, generators and drift detectors were implemented using Python 3.8 and the \texttt{river}~\cite{montiel2021river} package. Algorithms were run on a GNU/Linux cluster with 192 Intel Xeon cores, 6 TB RAM, and Centos 7.

\section{Results}
\label{sec:results}

This section presents and discusses the experimental results. Firstly, we conducted a comparative analysis of all drift detectors, irrespective of the specific scenarios, to determine which drift detector exhibits the best overall performance both with and without concept drift. Secondly, we investigate deeper into each scenario to gain insights into how the locality and the proposed data difficulties influence drift detection. Additionally, we evaluated the influence of the number of classes on drift detection in each scenario. Finally, we assessed how these proposed difficulties affect the classifier's performance and the impact of using a classifier in combination with an explicit drift detector.

\subsection{Overall aggregated comparison for all scenarios}
To address \textbf{RQ1}, we first evaluated the performance of all classifiers with a focus on data streams that either had or did not have concept drift, irrespective of how the concept drift affected the feature space or class space. Table~\ref{tab:stats_no_drift} presents the results for data streams without concept drift. In these cases, as the stream remains stationary, we display only the average True Negatives and False Positives values. Notably, only $3$ drift detectors did not raise any alert in this scenario, highlighting the sensitivity of explicit drift detectors to changes in the error distribution, even when data distribution remains stable. Moreover, ECDD, EDDM, and STEPD exhibited a high level of sensitivity to any change, resulting in a high number of False Positives. In contrast, DDM and ADWIN displayed the lowest values of false alerts, demonstrating their stability over time.

Table~\ref{tab:overall_metrics} presents the results considering data streams with concept drift. It is interesting to note that EDDM exhibited the lowest delay among all evaluated drift detectors and the second-highest recall, although this came at the cost of raising numerous drift alerts, leading to a precision of $0\%$. A similar pattern is observed for ECDD and STEPD. When comparing DDM and its variations, their detection performance is quite similar, with HDDM displaying the best F1-Score and the shortest detection delay, highlighting the efficiency of Hoeffding's inequality. In terms of F1-Score, ADWIN and PH displayed the best detection performance, with PH having a shorter detection delay.

\begin{table}[!b]
\centering
\caption{Comparison between drift detectors considering scenarios without concept drift. }
\label{tab:stats_no_drift}
\resizebox{.25\textwidth}{!}{%
\begin{tabular}{@{}lrr@{}}
\toprule
\textbf{Drift Detector} & \multicolumn{1}{l}{\textbf{TN}} & \multicolumn{1}{l}{\textbf{FP}} \\ \midrule
ADWIN & 0.00 & 3.79 \\
DDM & 0.17 & \textbf{2.71} \\
ECDD & \textbf{0.46} & 904.58 \\
EDDM & 0.00 & 62914.04 \\
HDDM & 0.00 & 41.83 \\
KSWIN & 0.08 & 10.25 \\
PH & 0.00 & 12.46 \\
RDDM & 0.00 & 27.96 \\
STEPD & 0.00 & 4628.13 \\ \bottomrule
\end{tabular}%
}
\end{table}

\begin{table}[!h]
\centering
\caption{Comparison between drift detectors considering all evaluated drift difficulties.}
\label{tab:overall_metrics}
\resizebox{.43\textwidth}{!}{%
\begin{tabular}{@{}lrrrr@{}}
\toprule
\textbf{Drift Detector} & \multicolumn{1}{l}{\textbf{Precision}} & \multicolumn{1}{l}{\textbf{Recall}} & \multicolumn{1}{l}{\textbf{F-1}} & \multicolumn{1}{l}{\textbf{Delay}} \\ \midrule
ADWIN & \textbf{7.40\%} & 34.50\% & \textbf{12.19\%} & 2192 \\
DDM & 1.82\% & 5.45\% & 2.73\% & 2168 \\
ECDD & 0.03\% & 36.48\% & 0.06\% & 235 \\
EDDM & 0.00\% & 95.18\% & 0.00\% & 15 \\
HDDM & 1.65\% & 76.61\% & 3.23\% & 1271 \\
KSWIN & 3.32\% & 42.54\% & 6.16\% & 1866 \\
PH & 5.18\% & 72.19\% & 9.66\% & 1867 \\
RDDM & 0.89\% & 25.99\% & 1.72\% & 2027 \\
STEPD & 0.01\% & \textbf{95.80\%} & 0.03\% & 516 \\ \bottomrule
\end{tabular}%
}
\end{table}

In summary, ADWIN, DDM, and PH stood out as strong performers in both evaluated scenarios. They exhibited caution when dealing with stationary streams and demonstrated good detection performance when concept drift occurred. On the other hand, EDDM, ECDD, and STEPD could detect concept drifts quickly but raised a significant number of false alarms in the process.

\subsection{Detailed comparison within each scenario}

Tables \ref{tab:metrics_single_class_local} and \ref{tab:metrics_single_class_global} present the metrics for Single-Class Local and Global concept drift scenarios. When we analyze the detection performance, it is evident that ADWIN and PH consistently exhibited the best detection performance for both local and global concept drifts, while EDDM, ECCD, and STEPD displayed the least favorable results. Comparing the differences in results between local and global concept drift, we observe that precision values remained relatively stable. However, recall values increased for most drift detectors, with a $17\%$ increase for ADWIN and a $10\%$ increase for PH. This indicates an inverse correlation between false alarms and the magnitude of the concept drift. On the other hand, the detection rate improved from local to global drifts. This is understandable since global concept drifts have a more substantial impact on data distribution, making them easier to detect due to their significant influence.

When considering Multi-Class concept drift scenarios, ADWIN and PH again demonstrated the best results for both scenarios. Similar to the observations in Single-Class concept drift, when we compare local and global drifts, we notice that global drifts were detected more effectively due to their more substantial impact on classifier performance. Furthermore, it is noteworthy that Multi-Class Local drifts exhibited higher recall values than their single-class counterparts. This is expected since a larger portion of the data distribution is influenced by the drift in multi-class scenarios, making it easier to detect and leading to improved recall values.

\begin{figure}[t!]
    \begin{subfigure}{\columnwidth}
    \centering
        \includegraphics[width=.6\textwidth]{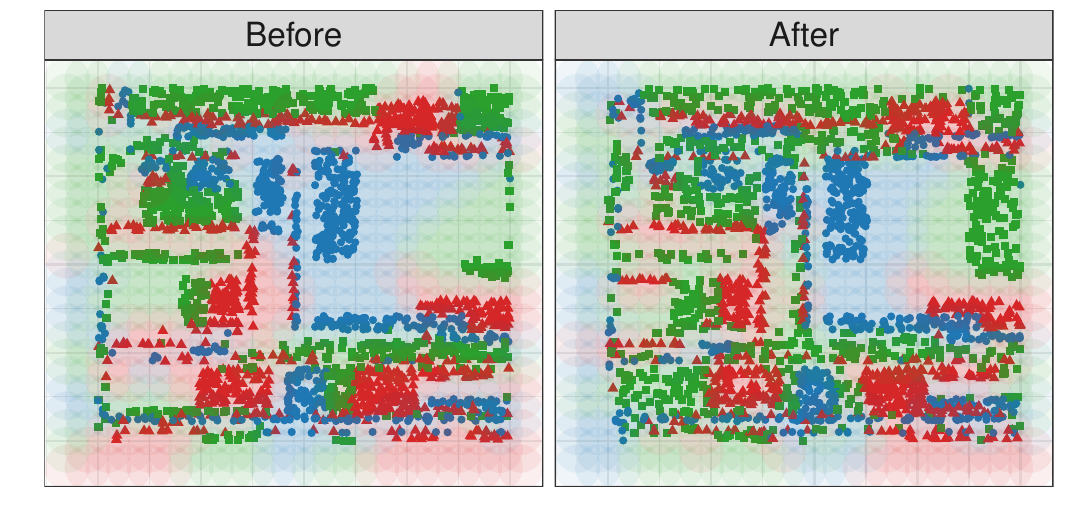}
        \caption{Data distribution before and after a concept drift where new boundaries for the green class are created.}
    \end{subfigure}
    \begin{subfigure}{\columnwidth}
        \centering
        \includegraphics[width=.75\textwidth]{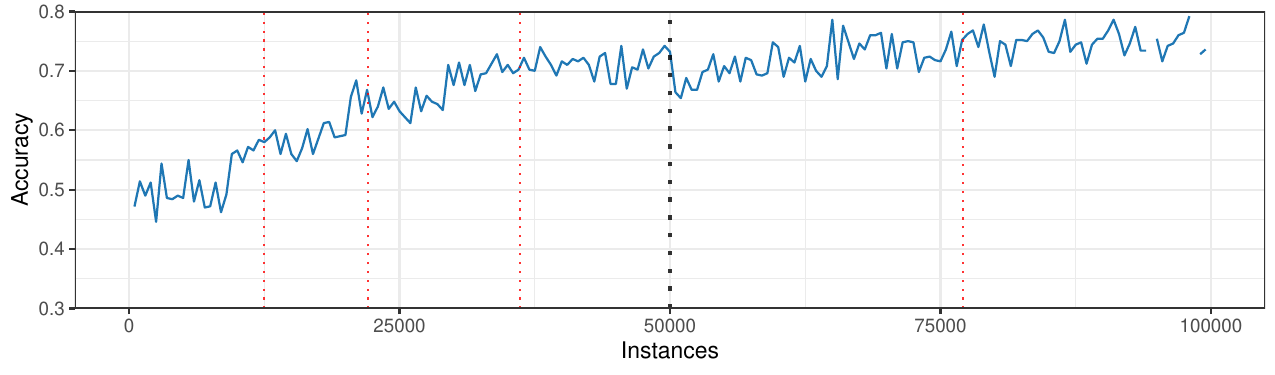}
        \caption{Hoeffing Tree accuracy over time for a specific stream experiencing concept drift. The gray rectangle denotes the duration of the drift. Red dashed lines represent the drift alarms triggered by ADWIN.}
    \end{subfigure}
    \caption{Single-Class Local sudden drift data distribution and drift alerts over time considering the \texttt{emerging\_branch} difficulty and ADWIN as drift detector.}
    \label{fig:single_local_acc_drift_alert}
\end{figure}

\begin{figure}[t!]
    \centering
    \begin{subfigure}{\columnwidth}
        \centering
        \includegraphics[width=.75\textwidth]{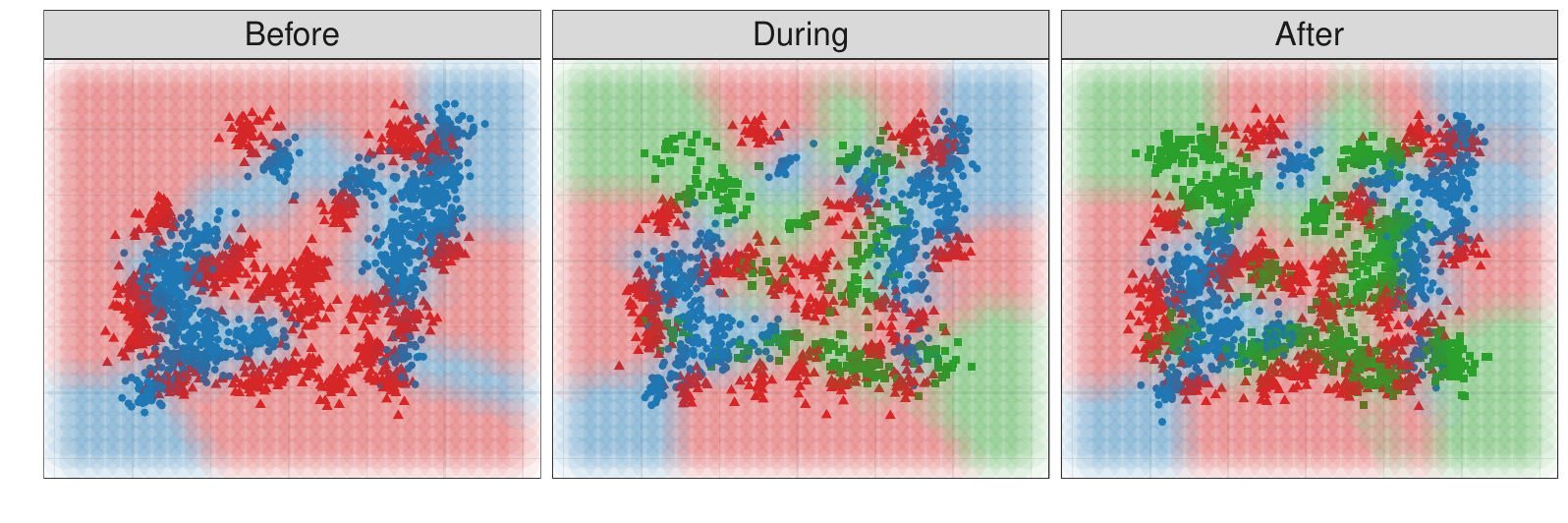}
        \caption{Data distribution before, during and after a concept drift where a new class (green) emerges.}
    \end{subfigure}
    \begin{subfigure}{\columnwidth}
        \centering
        \includegraphics[width=.75\textwidth]{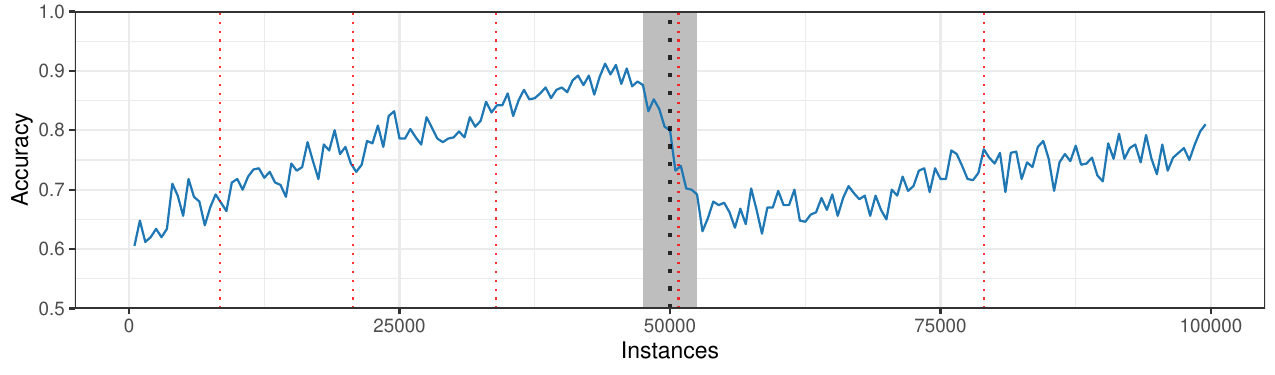}
        \caption{Hoeffing Tree accuracy over time for a specific stream experiencing concept drift. The gray rectangle denotes the duration of the drift. Red dashed lines represent the drift alarms triggered by ADWIN.}
    \end{subfigure}
    \caption{Single-Class Global drift data distribution and drift alerts over time considering the \texttt{class\_emerging} difficulty and ADWIN as drift detector.}
    \label{fig:single_global_acc_drift_alert}
    \vspace{.5cm}
\end{figure}

\begin{figure}[t!]
    \centering
    \begin{subfigure}[b]{\columnwidth}
        \centering
        \includegraphics[width=.6\textwidth]{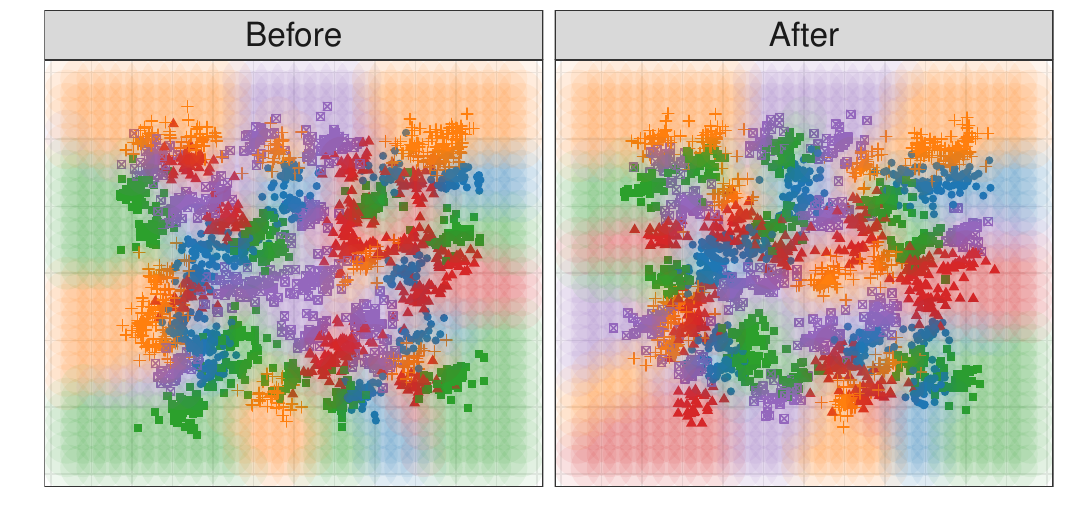}
        \caption{Data distribution before and after a concept drift that swaps positions of some clusters of all classes.}
    \end{subfigure}
    \begin{subfigure}[b]{\columnwidth}
        \centering
        \includegraphics[width=.75\textwidth]{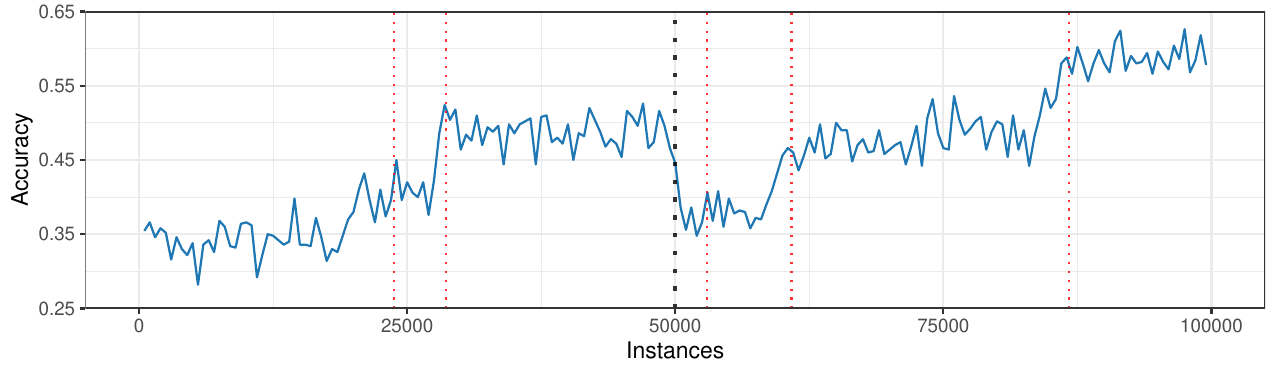}
        \caption{Hoeffing Tree accuracy over time for a specific stream experiencing concept drift. The gray rectangle denotes the duration of the drift. Red dashed lines represent the drift alarms triggered by ADWIN.}
    \end{subfigure}
    \caption{Multi-Class Local drift data distribution and drift alerts over time considering the \texttt{swap\_cluster} difficulty and ADWIN as drift detector.}
    \label{fig:multi_local_acc_drift_alert}
\end{figure}

\begin{figure}[h!]
    \centering
    \begin{subfigure}[b]{\columnwidth}
        \centering
        \includegraphics[width=.75\textwidth]{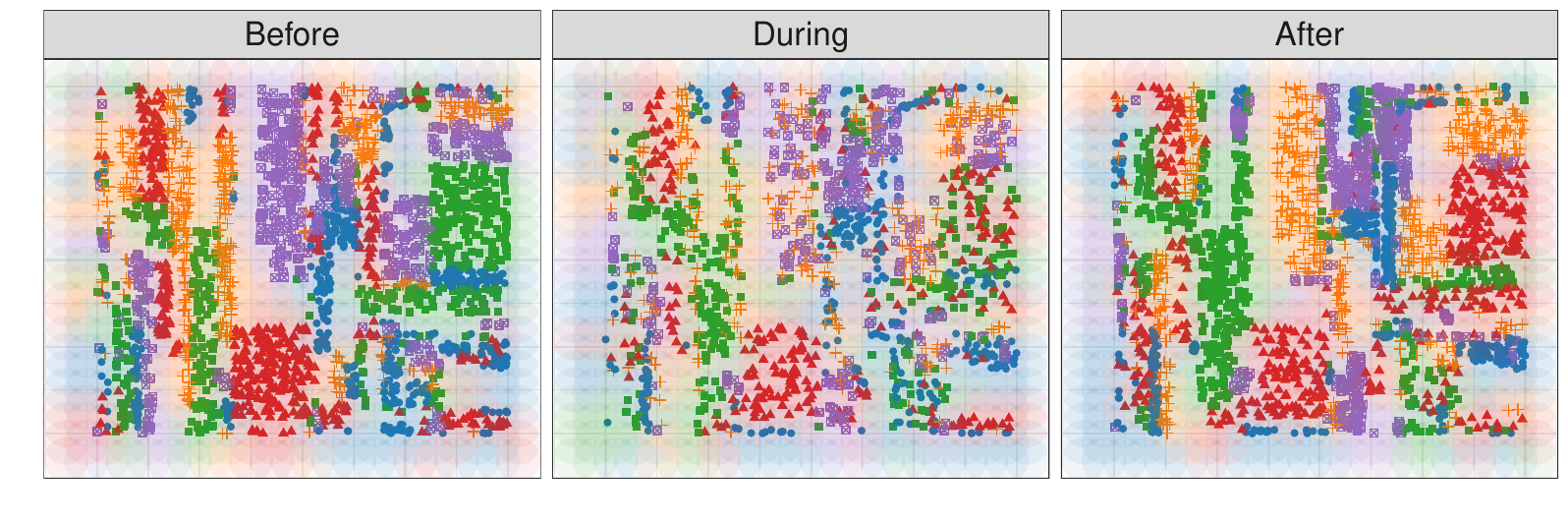}
        \caption{Data distribution before, during and after a concept drift that swaps positions of all classes.}
    \end{subfigure}
    \begin{subfigure}[b]{\columnwidth}
        \centering
        \includegraphics[width=.75\textwidth]{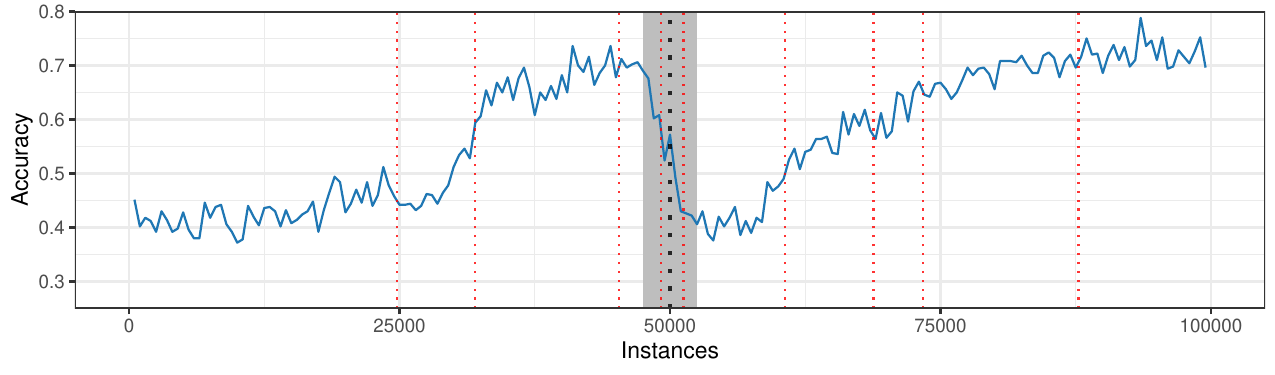}
        \caption{Hoeffing Tree accuracy over time for a specific stream experiencing concept drift. The gray rectangle denotes the duration of the drift. Red dashed lines represent the drift alarms triggered by ADWIN.}
    \end{subfigure}
    \caption{Multi-Class Global drift data distribution and drift alerts over time considering the \texttt{swap\_leaves} difficulty and ADWIN as drift detector.}
    \label{fig:multi_global_acc_drift_alert}
\end{figure}

Additionally, Figs.~\ref{fig:single_local_acc_drift_alert} to ~\ref{fig:multi_global_acc_drift_alert} illustrate four examples of the data distribution, accuracy of the classifier, and the moments when the ADWIN (the most effective overall) signaled drifts. These visualizations provide insight into the ease or difficulty in detecting specific scenarios. In the case of Local drifts (Figs.~\ref{fig:single_local_acc_drift_alert} and ~\ref{fig:multi_local_acc_drift_alert}), the drift alerts were signaled when no actual drift occurred, indicating that detecting Local drifts is challenging due to their subtle nature. Conversely, in Global drifts (Figs.~\ref{fig:single_global_acc_drift_alert} and ~\ref{fig:multi_global_acc_drift_alert}), a more distinct alteration in the data distribution was observable, making it easier to identify and leading to more noticeable changes in classifier accuracy. Both Single and Multi scenarios exhibited false alarms, where drift detection alerts were raised despite no actual drift occurring during those periods.  

To address \textbf{RQ2}, ADWIN and PH consistently demonstrated superior detection performance across all evaluated scenarios. Conversely, EDDM and STEPD, while achieving high recall values, also triggered numerous false alarms, resulting in lower precision. It is worth noting that DDM consistently exhibited low values of both recall and precision in all scenarios, indicating a more conservative approach. Furthermore, when comparing the behavior of drift detectors in the four scenarios, a hierarchy of difficulty emerges, from easiest to hardest detection as follows: Multi-Class Global, Single-Class Global, Multi-Class Local, Single-Class Local. This order of difficulty corresponds closely to the impact on data distribution, with a more pronounced impact leading to a more significant change in error distribution. Another noteworthy observation is that in all evaluated drift detectors, scenarios with more localized drifts tended to generate a higher number of false alarms. This highlights the need for the development of drift detectors that are less sensitive to error distribution, particularly in scenarios with local drifts. In conclusion, our analysis provides insights into how each scenario affected the performance of drift detectors, addressing RQ3.


\begin{table}[!h]
\centering
\caption{Comparison between drift detectors considering single class local concept drifts.}
\label{tab:metrics_single_class_local}
\resizebox{.43\textwidth}{!}{%
\begin{tabular}{@{}lrrrr@{}}
\toprule
\textbf{Drift Detector} & \multicolumn{1}{l}{\textbf{Precision}} & \multicolumn{1}{l}{\textbf{Recall}} & \multicolumn{1}{l}{\textbf{F1}} & \multicolumn{1}{l}{\textbf{Delay}} \\ \midrule
ADWIN & \textbf{5.48\%} & 22.40\% & 8.80\% & 2728 \\
DDM & 1.88\% & 5.47\% & 2.80\% & 2287 \\
ECDD & 0.03\% & 29.17\% & 0.07\% & 201 \\
EDDM & 0.00\% & \textbf{97.40\%} & 0.00\% & 4 \\
HDDM & 1.69\% & 67.45\% & 3.30\% & 1359 \\
KSWIN & 2.91\% & 31.51\% & 5.32\% & 1991 \\
PH & 5.21\% & 64.06\% & \textbf{9.63\%} & 2176 \\
RDDM & 1.13\% & 31.77\% & 2.18\% & 2016 \\
STEPD & 0.02\% & 91.15\% & 0.04\% & 649 \\ \bottomrule
\end{tabular}%
}
\end{table}

\begin{table}[!h]
\centering
\caption{Comparison between drift detectors considering single class global concept drifts.}
\label{tab:metrics_single_class_global}
\resizebox{.43\textwidth}{!}{%
\begin{tabular}{@{}lrrrr@{}}
\toprule
\textbf{Drift Detector} & \multicolumn{1}{l}{\textbf{Precision}} & \multicolumn{1}{l}{\textbf{Recall}} & \multicolumn{1}{l}{\textbf{F1}} & \multicolumn{1}{l}{\textbf{Delay}} \\ \midrule
ADWIN & \textbf{8.15\%} & 39.06\% & \textbf{13.48\%} & 1909 \\
DDM & 2.21\% & 6.25\% & 3.27\% & 2286 \\
ECDD & 0.02\% & 26.30\% & 0.05\% & 315 \\
EDDM & 0.00\% & \textbf{91.41\%} & 0.00\% & 11 \\
HDDM & 1.81\% & 69.01\% & 3.52\% & 1321 \\
KSWIN & 3.30\% & 34.11\% & 6.02\% & 1879 \\
PH & 5.87\% & 75.78\% & 10.90\% & 1737 \\
RDDM & 1.01\% & 28.39\% & 1.96\% & 1982 \\
STEPD & 0.01\% & 93.75\% & 0.02\% & 517 \\ \bottomrule
\end{tabular}%
}
\end{table}

\begin{table}[!h]
\centering
\caption{Comparison between drift detectors considering multi class local concept drifts.}
\label{tab:metrics_multi_class_local}
\resizebox{.43\textwidth}{!}{%
\begin{tabular}{@{}lrrrr@{}}
\toprule
\textbf{Drift Detector} & \multicolumn{1}{l}{\textbf{Precision}} & \multicolumn{1}{l}{\textbf{Recall}} & \multicolumn{1}{l}{\textbf{F1}} & \multicolumn{1}{l}{\textbf{Delay}} \\ \midrule
ADWIN & \textbf{7.19\%} & 30.40\% & \textbf{11.63\%} & 2360 \\
DDM & 1.58\% & 4.67\% & 2.36\% & 2036 \\
ECDD & 0.04\% & 40.29\% & 0.07\% & 192 \\
EDDM & 0.00\% & \textbf{95.51\%} & 0.00\% & 21 \\
HDDM & 1.67\% & 80.68\% & 3.28\% & 1207 \\
KSWIN & 3.31\% & 44.32\% & 6.16\% & 1873 \\
PH & 5.18\% & 70.88\% & 9.66\% & 1956 \\
RDDM & 0.92\% & 26.37\% & 1.77\% & 2083 \\
STEPD & 0.02\% & 96.70\% & 0.03\% & 508 \\ \bottomrule
\end{tabular}%
}
\end{table}

\begin{table}[t!]
\centering
\caption{Comparison between drift detectors considering multi class global concept drifts.}
\label{tab:metrics_multi_class_global}
\resizebox{.43\textwidth}{!}{%
\begin{tabular}{@{}lrrrr@{}}
\toprule
\textbf{Drift Detector} & \multicolumn{1}{l}{\textbf{Precision}} & \multicolumn{1}{l}{\textbf{Recall}} & \multicolumn{1}{l}{\textbf{F1}} & \multicolumn{1}{l}{\textbf{Delay}} \\ \midrule
ADWIN & \textbf{7.95\%} & 42.92\% & \textbf{13.41\%} & 2033 \\
DDM & 1.94\% & 6.05\% & 2.93\% & 2195 \\
ECDD & 0.03\% & 39.38\% & 0.06\% & 277 \\
EDDM & 0.00\% & \textbf{95.43\%} & 0.00\% & 14 \\
HDDM & 1.56\% & 78.88\% & 3.06\% & 1299 \\
KSWIN & 3.48\% & 48.86\% & 6.49\% & 1819 \\
PH & 4.91\% & 75.80\% & 9.22\% & 1706 \\
RDDM & 0.72\% & 21.92\% & 1.39\% & 1977 \\
STEPD & 0.01\% & 97.60\% & 0.03\% & 469 \\ \bottomrule
\end{tabular}%
}
\end{table}

\subsection{Comparison between difficulties}

In this experiment, our aim is to assess the varying levels of difficulty encountered by each drift detector when dealing with different concept drift scenarios. We seek to analyze which scenarios pose more significant challenges and which ones are relatively easier to detect for each drift detector. To accommodate space limitations within the manuscript, we will primarily focus on presenting the results for the top-performing drift detectors, ADWIN and Page Hinkley. The complete set of results is accessible on our website for a more detailed examination.

When considering Single-Class Local drifts, Table~\ref{tab:single_class_local_difficulties} provides an overview of the performance metrics for each difficulty as observed with both drift detectors. It is worth noting that, for both detectors, the \texttt{reappering\_cluster} scenario appeared to be the easiest to detect. This is because the disappearance and reappearance of data segments tend to induce substantial changes in error distribution, making them relatively clear for the drift detectors, even if they occur in specific regions of the feature space. Conversely, when faced with more complex drift scenarios, ADWIN displayed its worst performance in detecting the \texttt{emerging\_branch} difficulty, while PH struggled with the \texttt{moving\_cluster} scenario. The differences in their performance can be attributed to their distinct detection mechanisms. ADWIN employs a two-window approach that relies on detecting a significant difference between the two windows to identify concept drift. Consequently, when the classifier rapidly adapts to the new data distribution, ADWIN may encounter challenges in detecting the drift. On the other hand, PH utilizes a cumulative approach, allowing it to handle such scenarios more effectively. Additionally, difficulties that inherently involve incremental changes, such as those related to moving, merging, and splitting clusters, displayed a higher detection delay in general.


Furthermore, Table~\ref{tab:single_class_global_difficulties} provides an overview of the results for each of the Single-Class Global difficulties. Notably, for both ADWIN and PH, the \texttt{class\_emerging} difficulty appeared to be the least challenging to detect and exhibited the lowest detection delay. This observation aligns with the logic that the emergence of a new class results in a sudden drop in accuracy, as a novel class is being introduced and evaluated. Conversely, the most challenging difficulty to detect was the \texttt{moving\_cluster}, followed by \texttt{splitting\_cluster} and \texttt{merging\_cluster}. These three difficulties are inherently incremental in nature, which means they do not create an abrupt change in the learning curve and error rate. Consequently, this incremental nature makes them harder to detect promptly. Comparing the results to those of Single-Class Local drifts, an improvement in recall values is noticeable. This suggests that, for almost all difficulties, detecting Global drifts was generally easier due to their broader impact on the data distribution.

Analyzing the metrics for Multi-Class Local drifts in Table~\ref{tab:multi_class_local_difficulties}, ADWIN achieved the best results when encountering the \texttt{prune\_growth\_new\_branch} difficulty, while PH exhibited its best performance with \texttt{swap\_clusters}. Notably, results for difficulties that do not involve incremental drifts were quite similar to the best results. This observation emphasizes the challenge that drift detectors face in effectively detecting incremental drifts, even when multiple classes are affected. Another aspect worth noting is the larger performance gap between ADWIN and PH in the context of multi-class drifts compared to single-class drifts. This difference can be attributed to the higher precision that ADWIN demonstrates in multi-class drift scenarios. When we make comparisons with their Single-Class Local counterparts, it becomes evident that multi-class drifts generally yield better detection results, as expected due to their greater impact on data distribution.

The results for Multi-Class Global drifts are presented in Table~\ref{tab:multi_class_global_difficulties}. ADWIN displayed its best detection performance when confronted with \texttt{prune\_g\-rowth\_new\_branch} drifts, but it also achieved the satisfactory F1 values for other difficulties, such as \texttt{prune\_regrowth\_branch} and \texttt{swap\_leaves}. The poorest results for ADWIN once again correlated with difficulties involving incremental drift. PH, on the other hand, exhibited its best performance with \texttt{reappearing\_cluster}, although its performance with other difficulties was quite similar and generally less favorable than that of ADWIN. In comparison to Multi-Class Local drifts, precision values remained similar, but recall values increased, indicating improved detection even with a notable number of false alarms.

To address \textbf{RQ4}, it becomes evident that sudden changes in data distribution, like new data appearing in the feature space or the emergence of a new class, were the easiest and quickest to be detected. Conversely, difficulties that inherently involved incremental changes displayed higher values of False Negatives, resulting in lower recall and less effective detection. It is important to note that this observation is correlated with how the base learner operates. Consequently, when evaluating drift detectors with other base learners or employing unsupervised drift detectors, the ranking of difficulties in terms of complexity may vary.

\begin{table}[b!]
\centering
\caption{Comparison of detection performance for Single-Class Local drift difficulties by ADWIN and Page Hinkley. Bold values represent the highest value of that column (easiest detection) and underscored values represent the lowest value (harder detection).}
\label{tab:single_class_local_difficulties}
\resizebox{.8\textwidth}{!}{%
\begin{tabular}{@{}lcccccccc@{}}
\toprule
\multirow{2}{*}{\textbf{Difficulty}} & \multicolumn{2}{c}{\textbf{Precision}} & \multicolumn{2}{c}{\textbf{Recall}} & \multicolumn{2}{c}{\textbf{F1}} & \multicolumn{2}{c}{\textbf{Delay}} \\ \cmidrule(l){2-9} 
 & ADWIN & PH & ADWIN & PH & ADWIN & PH & ADWIN & PH \\ \midrule
emerging\_branch & \underline{3.83\%} & 5.24\% & \underline{14.58\%} & \textbf{70.83\%} & \underline{6.06\%} & 9.76\% & 2176 & 1975 \\
emerging\_cluster & 5.61\% & 5.41\% & 25.00\% & 62.50\% & 9.16\% & 9.95\% & 3364 & 2249 \\
merging\_cluster & 5.15\% & 5.39\% & 20.83\% & 62.50\% & 8.26\% & 9.92\% & 2914 & 2285 \\
moving\_cluster & 5.03\% & \underline{4.94\%} & 20.83\% & \underline{58.33\%} & 8.10\% & \underline{9.11\%} & 2930 & 2050 \\
prune\_growth\_new\_branch & 4.47\% & 5.09\% & 16.67\% & 66.67\% & 7.05\% & 9.45\% & 2257 & 2067 \\
prune\_regrowth\_branch & 7.14\% & 4.94\% & 22.92\% & 64.58\% & 10.89\% & 9.17\% & 2479 & 2261 \\
reappearing\_cluster & \textbf{7.35\%} & \textbf{5.56\%} & \textbf{37.50\%} & 66.67\% & \textbf{12.29\%} & \textbf{10.26\%} & 2550 & 2179 \\
splitting\_cluster & 4.95\% & 5.16\% & 20.83\% & 60.42\% & 8.00\% & 9.51\% & 2936 & 2368 \\ \bottomrule
\end{tabular}%
}
\end{table}

\begin{table}[b!]
\centering
\caption{Comparison of detection performance for Single-Class Global drift difficulties by ADWIN and Page Hinkley. Bold values represent the highest value of that column (easiest detection) and underscored values represent the lowest value (harder detection).}
\label{tab:single_class_global_difficulties}
\resizebox{.8\textwidth}{!}{%
\begin{tabular}{@{}lcccccccc@{}}
\toprule
\multirow{2}{*}{\textbf{Difficulty}} & \multicolumn{2}{c}{\textbf{Precision}} & \multicolumn{2}{c}{\textbf{Recall}} & \multicolumn{2}{c}{\textbf{F1}} & \multicolumn{2}{c}{\textbf{Delay}} \\ \cmidrule(l){2-9} 
 & ADWIN & PH & ADWIN & PH & ADWIN & PH & ADWIN & PH \\ \midrule
class\_emerging\_rbf & 10.07\% & \textbf{7.68\%} & \textbf{58.33\%} & \textbf{93.75\%} & 17.18\% & \textbf{14.20\%} & 1104 & 1698 \\
class\_emerging\_rt & \textbf{11.06\%} & 6.55\% & 52.08\% & 85.42\% & \textbf{18.25\%} & 12.17\% & 1198 & 1280 \\
merging\_cluster & 5.24\% & 5.17\% & 20.83\% & 60.42\% & 8.37\% & 9.52\% & 2895 & 2130 \\
moving\_cluster & \underline{4.37\%} & \underline{4.32\%} & 20.83\% & \underline{54.17\%} & \underline{7.22\%} & \underline{8.00\%} & 2885 & 2089 \\
prune\_growth\_new\_branch & 9.60\% & 5.68\% & 39.58\% & 77.08\% & 15.45\% & 10.59\% & 2378 & 1689 \\
prune\_regrowth\_branch & 9.21\% & 5.54\% & 43.75\% & 83.33\% & 15.22\% & 10.39\% & 1722 & 1487 \\
reappearing\_cluster & 9.62\% & 6.76\% & 58.33\% & 91.67\% & 16.52\% & 12.59\% & 2124 & 1896 \\
splitting\_cluster & 4.50\% & 5.22\% & \underline{18.75\%} & 60.42\% & 7.26\% & 9.60\% & 2985 & 1899 \\ \bottomrule
\end{tabular}%
}
\end{table}

\begin{table}[t!]
\centering
\caption{Comparison of detection performance for Multi-Class Local drift difficulties by ADWIN and Page Hinkley. Bold values represent the highest value of that column (easiest detection) and underscored values represent the lowest value (harder detection).}
\label{tab:multi_class_local_difficulties}
\resizebox{.8\textwidth}{!}{%
\begin{tabular}{@{}lcccccccc@{}}
\toprule
\multirow{2}{*}{\textbf{Difficulty}} & \multicolumn{2}{c}{\textbf{Precision}} & \multicolumn{2}{c}{\textbf{Recall}} & \multicolumn{2}{c}{\textbf{F1}} & \multicolumn{2}{c}{\textbf{Delay}} \\ \cmidrule(l){2-9} 
 & ADWIN & PH & ADWIN & PH & ADWIN & PH & ADWIN & PH \\ \midrule
emerging\_branch & \textbf{11.53\%} & 5.35\% & 45.37\% & 76.85\% & 18.39\% & 10.01\% & 2474 & 1644 \\
emerging\_cluster & 6.82\% & 5.15\% & 27.78\% & 64.81\% & 10.95\% & 9.54\% & 3016 & 2329 \\
merging\_cluster & 4.65\% & 5.14\% & 19.44\% & 64.81\% & 7.50\% & 9.52\% & 2783 & 2032 \\
moving\_cluster & 4.66\% & 4.86\% & 20.37\% & 63.89\% & 7.59\% & 9.03\% & 2799 & 2042 \\
prune\_growth\_new\_branch & 11.40\% & 5.37\% & \textbf{49.07\%} & 80.56\% & \textbf{18.50\%} & 10.08\% & 1983 & 1618 \\
prune\_regrowth\_branch & 4.37\% & \underline{3.88\%} & 15.00\% & \underline{58.33\%} & 6.77\% & \underline{7.27\%} & 2650 & 1970 \\
reappearing\_cluster & 11.37\% & 5.69\% & 48.33\% & 73.33\% & 18.41\% & 10.56\% & 2696 & 1749 \\
split\_node & \underline{3.32\%} & 4.50\% & \underline{12.04\%} & 61.11\% & \underline{5.21\%} & 8.38\% & 1777 & 2387 \\
splitting\_cluster & 4.82\% & 5.58\% & 20.37\% & 67.59\% & 7.80\% & 10.31\% & 3013 & 2213 \\
swap\_cluster & 7.76\% & \textbf{5.97\%} & 42.59\% & \textbf{85.19\%} & 13.12\% & \textbf{11.15\%} & 1995 & 1842 \\
swap\_leaves & 8.23\% & 5.23\% & 35.19\% & 78.70\% & 13.33\% & 9.80\% & 1675 & 1835 \\ \bottomrule
\end{tabular}%
}
\end{table}

\begin{table}[t!]
\centering
\caption{Comparison of detection performance for Multi-Class Global drift difficulties by ADWIN and Page Hinkley. Bold values represent the highest value of that column (easiest detection) and underscored values represent the lowest value (harder detection).}
\label{tab:multi_class_global_difficulties}
\resizebox{.75\textwidth}{!}{%
\begin{tabular}{@{}lcccccccc@{}}
\toprule
\multirow{2}{*}{\textbf{Difficulty}} & \multicolumn{2}{c}{\textbf{Precision}} & \multicolumn{2}{c}{\textbf{Recall}} & \multicolumn{2}{c}{\textbf{F1}} & \multicolumn{2}{c}{\textbf{Delay}} \\ \cmidrule(l){2-9} 
 & ADWIN & PH & ADWIN & PH & ADWIN & PH & ADWIN & PH \\ \midrule
merging\_cluster & \underline{4.37\%} & 4.53\% & \underline{20.37\%} & 60.19\% & \underline{7.20\%} & 8.43\% & 2917 & 1889 \\
moving\_cluster & 4.93\% & \underline{4.30\%} & 27.78\% & 66.67\% & 8.37\% & \underline{8.08\%} & 3148 & 1885 \\
prune\_growth\_new\_branch & \textbf{11.52\%} & 5.19\% & 63.89\% & 87.04\% & \textbf{19.52\%} & 9.80\% & 1666 & 1540 \\
prune\_regrowth\_branch & 10.19\% & 5.14\% & 55.00\% & 90.00\% & 17.19\% & 9.72\% & 1865 & 1485 \\
reappearing\_cluster & 9.31\% & \textbf{5.51\%} & 58.33\% & 85.00\% & 16.06\% & \textbf{10.34\%} & 1797 & 1705 \\
split\_node & 7.24\% & 4.79\% & 25.93\% & 67.59\% & 11.31\% & 8.94\% & 2129 & 2061 \\
splitting\_cluster & 4.97\% & 4.62\% & 20.37\% & \underline{58.33\%} & 7.99\% & 8.55\% & 2987 & 2177 \\
swap\_cluster & 8.76\% & 5.12\% & \textbf{69.44\%} & \textbf{91.67\%} & 15.56\% & 9.70\% & 1847 & 1263 \\
swap\_leaves & 9.79\% & 5.13\% & 57.41\% & 86.11\% & 16.73\% & 9.68\% & 1657 & 1608 \\ \bottomrule
\end{tabular}%
}
\end{table}

\subsection{Impact of number of classes and number of features}

\begin{figure}[b!]
    \centering
    \includegraphics[width=.7\columnwidth]{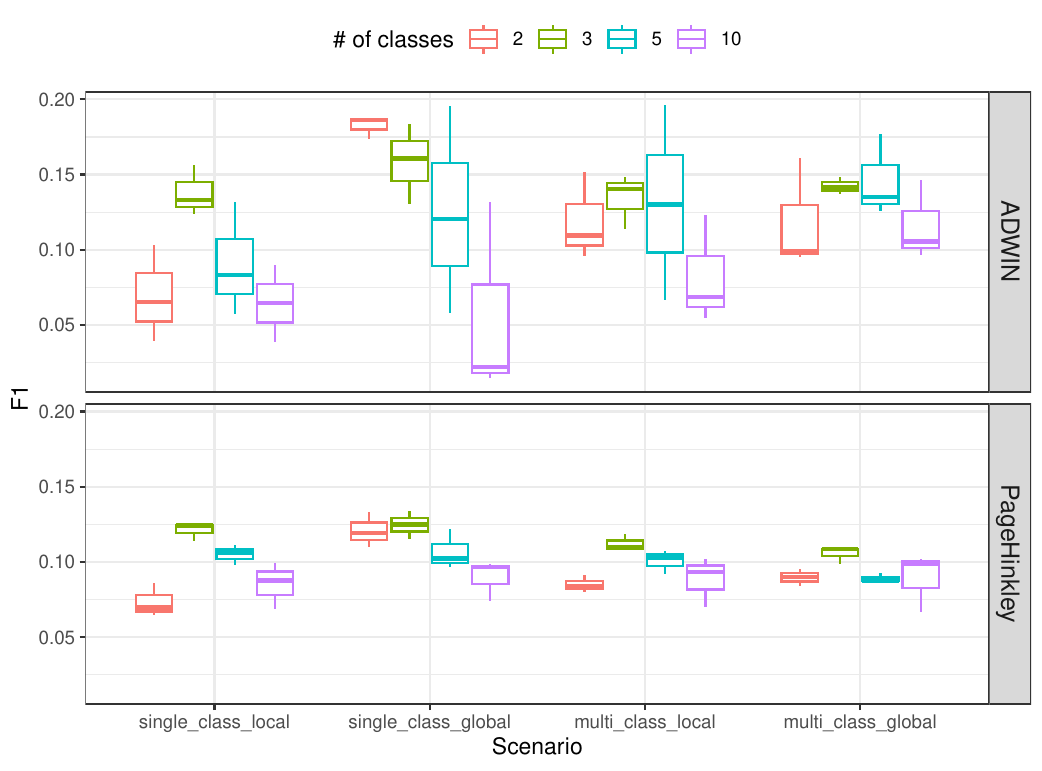}
    \caption{F1 Score for each categorization of concept drift. Each color represents streams with different number of classes.}
    \label{fig:n_classes_boxplot}
\end{figure}

\begin{figure}[t!]
    \centering
    \includegraphics[width=.7\columnwidth]{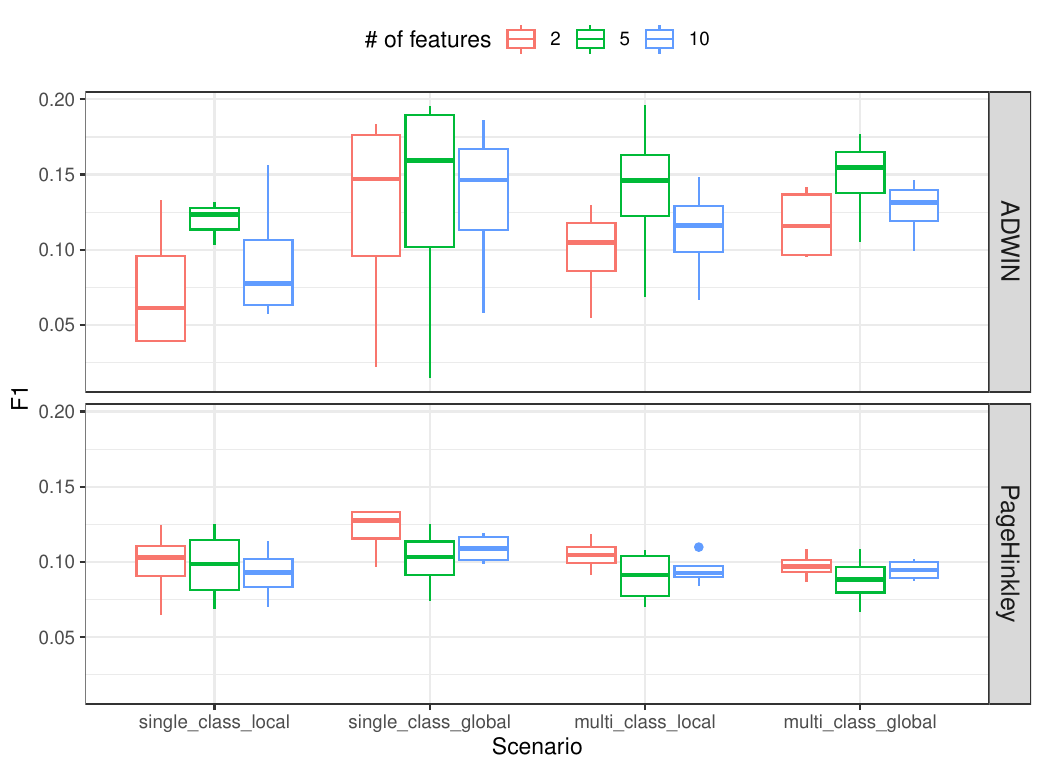}
    \caption{F1 Score for each categorization of concept drift. Each color represents streams with different number of features.}
    \label{fig:n_features_boxplot}
\end{figure}

Furthermore, the benchmark dataset used in the experiments included streams with varying configurations in terms of the number of classes and features. Therefore, we assessed how these different specifications influenced drift detection while considering each drift categorization to address \textbf{RQ5}. Fig.~\ref{fig:n_classes_boxplot} presents a boxplot showing the F1 score for ADWIN and Page Hinkley in each drift category, considering different numbers of classes, and Fig.~\ref{fig:n_features_boxplot} considering different feature space dimensionality. 

Firstly, it is evident that as the number of classes in the stream increases, the detection of concept drifts becomes more challenging, regardless of how the stream is affected. Moreover, when considering Single-Class Global drifts, a noticeable drop in performance is observed. This is expected in this specific scenario because when only one class is affected, the more stationary the other classes are, the more localized the change becomes, making it difficult to detect. On the other hand, when multiple classes are affected globally, the increase in the number of classes did not have a significant impact on drift detection. The scenario that posed the most challenge for both drift detectors was Single-Class Local drift with only $2$ classes. In a binary setting where only a portion of the classes is affected, it is similar to an imbalanced scenario, and these changes may not significantly impact the classifier's error.

When considering different feature space configurations, we can see that this factor did not have a substantial impact on Page Hinkley's performance. However, it did affect ADWIN, particularly in the presence of Single-Class Local and Multi-Class Local drifts. Additionally, it is worth noting that ADWIN's performance exhibits higher variance, while Page Hinkley demonstrates consistent performance across various stream configurations.

\clearpage
\subsection{Impact on classifier performance}

Finally, in addition to assessing how the proposed scenarios and difficulties influence concept drift detection, we also examined their impact on the performance of the utilized classifier, the Hoeffing Tree, when used in conjunction with the best-performing drift detector. The primary objective of this experiment is to ascertain how each of the proposed scenarios affects classification performance and to determine whether combining them with a drift detector enhances overall performance.

\begin{figure}[b!]
    \vspace{1.5cm}
    \centering
    \begin{subfigure}[b]{\columnwidth}
        \centering
        \includegraphics[width=.75\columnwidth]{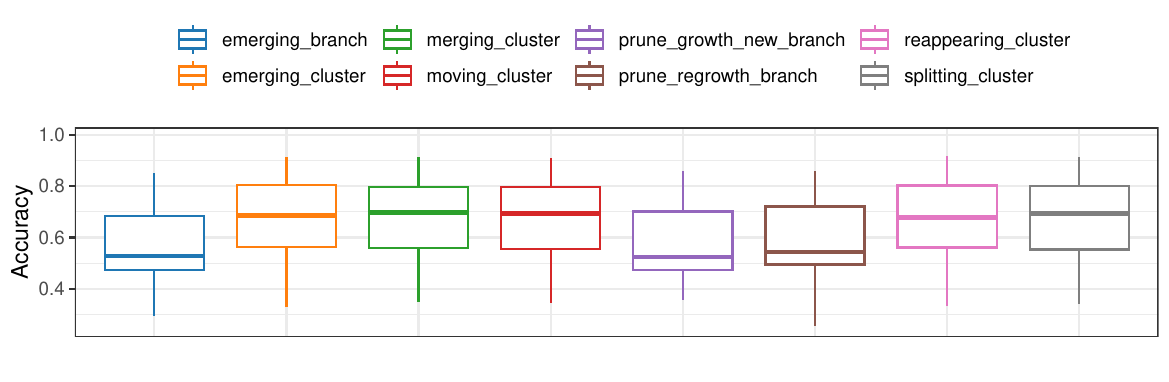}
        \caption{Single-Class Local}
    \end{subfigure}
    \begin{subfigure}[b]{\columnwidth}
        \centering
        \includegraphics[width=.75\columnwidth]{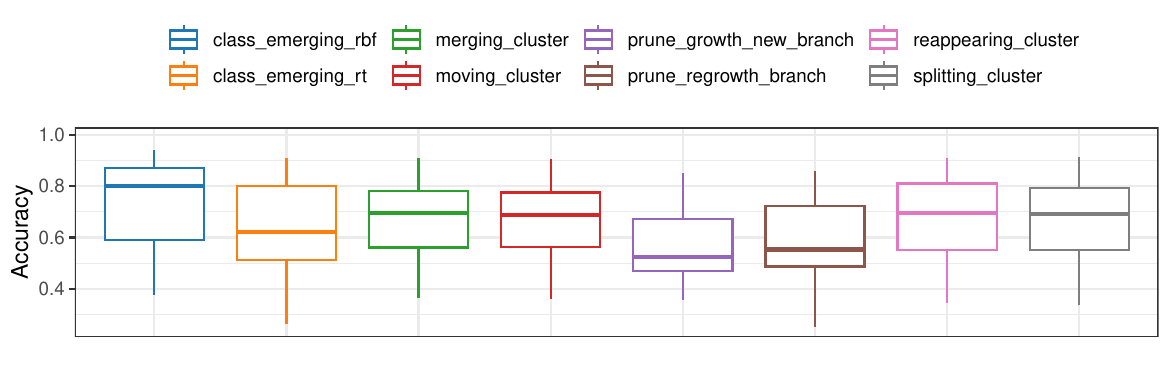}
        \caption{Single-Class Global}
    \end{subfigure}

    \caption{Boxplot of HT accuracy for each of the proposed difficulties for Single-Class Local (a) and Global (b)}
    \label{fig:accuracy_single_class}
\end{figure}

\begin{figure}[t!]
    \centering
    \begin{subfigure}[b]{.75\columnwidth}
        \centering
        \includegraphics[width=\columnwidth]{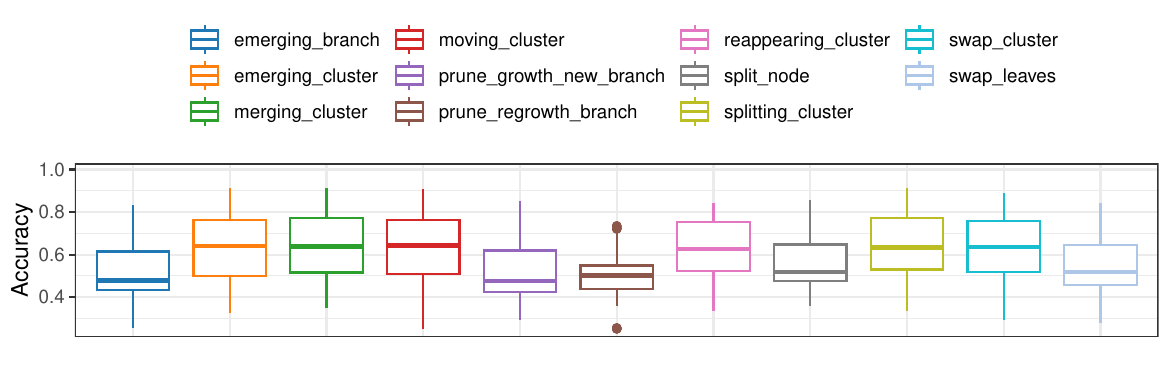}
        \caption{Multi-Class Local}
    \end{subfigure}
    \begin{subfigure}[b]{.75\columnwidth}
        \centering
        \includegraphics[width=\columnwidth]{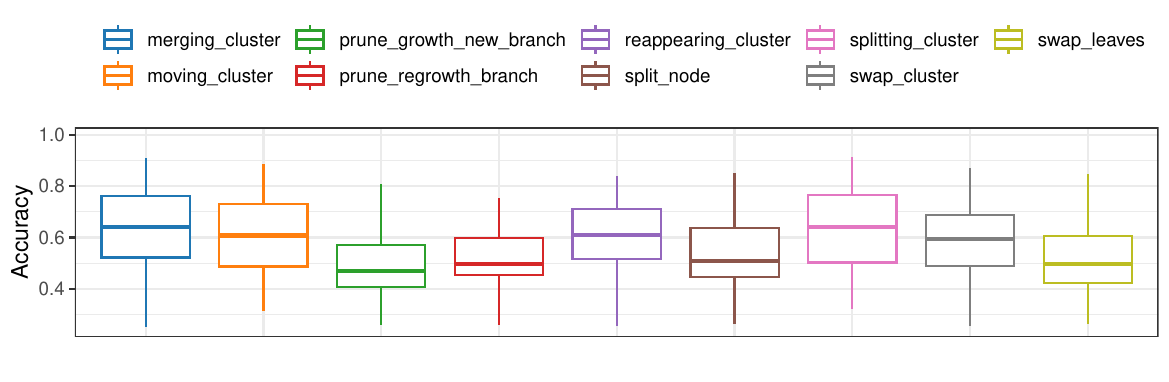}
        \caption{Multi-Class Global}
    \end{subfigure}

    \caption{Boxplot of HT accuracy for each of the proposed difficulties for Multi-Class Local (top) and Global (bottom)}
    \label{fig:accuracy_multi_class}
\end{figure}

Figs.~\ref{fig:accuracy_single_class} and \ref{fig:accuracy_multi_class} present the accuracy corresponding to each drift difficulty. Analyzing the average accuracy enables identification of the drifts causing the most significant drop in accuracy, given that, prior to the drift, all data streams from the same generator are equal.

In the Single-Class scenario, particularly focusing on local drifts, incremental changes such as emerging, merging, moving, and splitting clusters proved to be the most challenging to detect. However, classifiers exhibited a capacity for self-adaptation and showed their best accuracy levels. The emergence of new branches posed the most complex scenario in terms of predictive performance. Furthermore, streams generated with Random Tree generator displayed lower accuracy compared to those with RBF. In terms of global changes, the results were similar to local changes, except for \texttt{class\_emerging}, which exhibited higher accuracy values. This is likely due to the reduction in the number of classes for the majority of the stream, leading to improved accuracy.

Within the Multi-Class local and global scenarios, performance aligned closely with what was observed in Single-Class drifts. Incremental drifts demonstrated the highest accuracy levels, whereas \texttt{emerging\_branch} and \texttt{swap\_leaves} presented the most challenging difficulties. Once again, the Random Tree generator proved more challenging to learn from than RBF, which could be attributed to the distribution of clusters. As feature space dimensions increased, the RBF model facilitated easier learning of class boundaries.

Furthermore, we conducted a comparison to understand the impact of integrating the best-performing drift detector, i.e. ADWIN, on the classifier's performance. For this analysis, we compared the standard classifier with two modified versions: Drift/Warning Retraining (HT-DW) and Adaptive Hoeffding Tree (AHT). The former initiates retraining of the classifier upon the detector signaling a warning and completely replaces the classifier when a drift is detected. In contrast, the latter employs a drift detector for each branch of the tree, replacing only the affected branch when a drift signal is detected, while maintaining the overall structure of the tree. This differentiates HT-DW's global approach from AHT's localized strategy in addressing concept drift within the classifier structure.

Table~\ref{tab:acc_drift} displays the average accuracy for each drift category across the three classifiers. Overall, the AHT strategy, focusing on addressing drift locally, improved the accuracy of the standard Hoeffding Tree by an average of $4.03\%$. This improvement underscores the robustness of the Hoeffding Tree in adapting to new concepts, exhibiting good accuracy values even without a drift detector. On the other hadnd, despite the utilization of a drift detector, completely retraining the classifier resulted in decreased accuracy across all evaluated scenarios. Notably, the more localized the drift category, the more pronounced the reduction in accuracy.

\begin{figure}[h!]
    \centering
    \includegraphics[width=\textwidth]{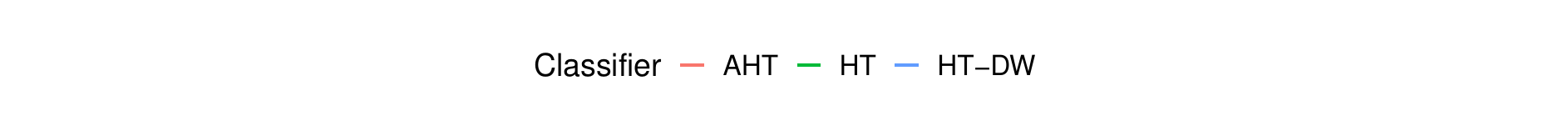}
    \includegraphics[width=.24\textwidth]{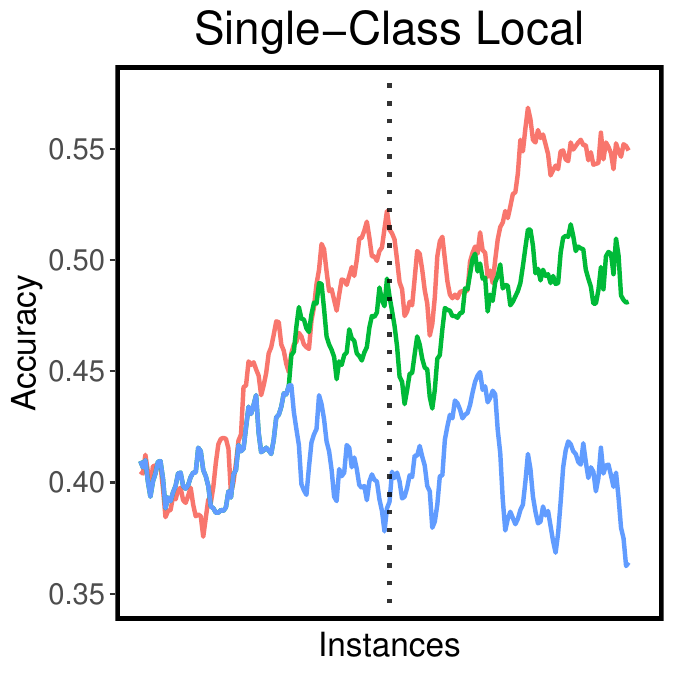}
    \includegraphics[width=.24\textwidth]{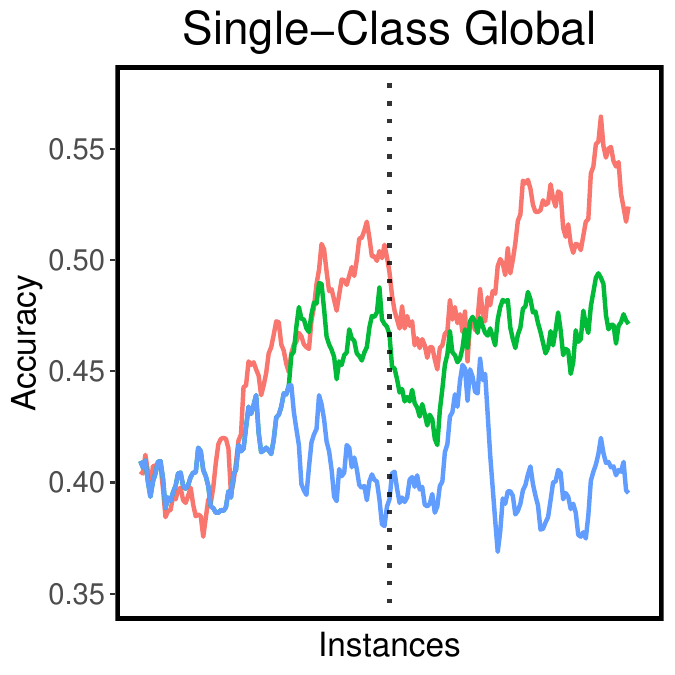}
    \includegraphics[width=.24\textwidth]{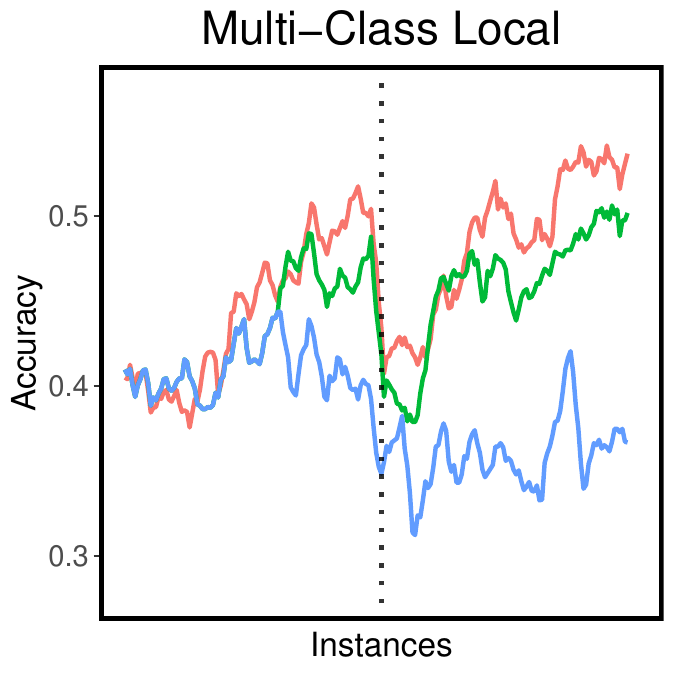}
    \includegraphics[width=.24\textwidth]{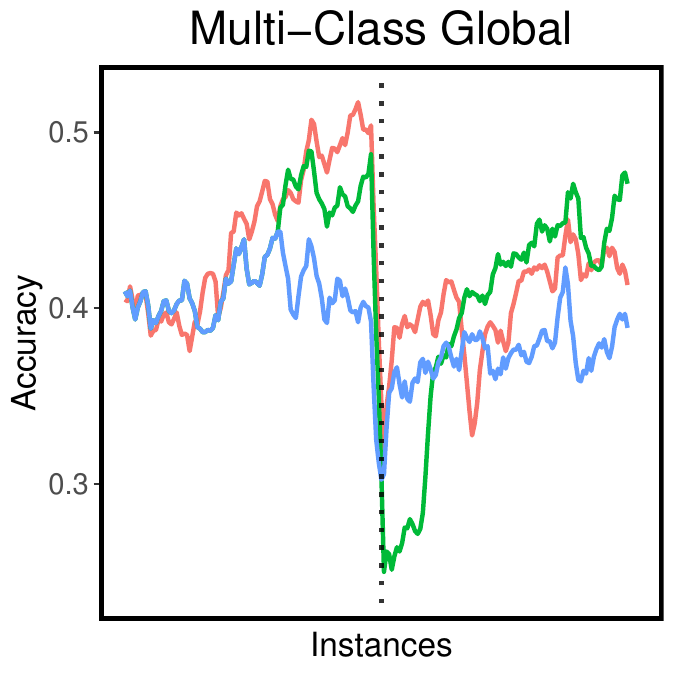}
    \caption{Accuracy curve for AHT, HT and HT-DW on \texttt{prune\_growth\_new\_branch} data stream in the presence of a sudden concept drift. Dashed vertical line indicates where the concept drift happened.}
    \label{fig:acc_prune_overall}

\end{figure}

\begin{figure}[t!]
    \vspace{-1cm}
    \centering
    \includegraphics[width=\textwidth]{legend.pdf}
    \includegraphics[width=.24\textwidth]{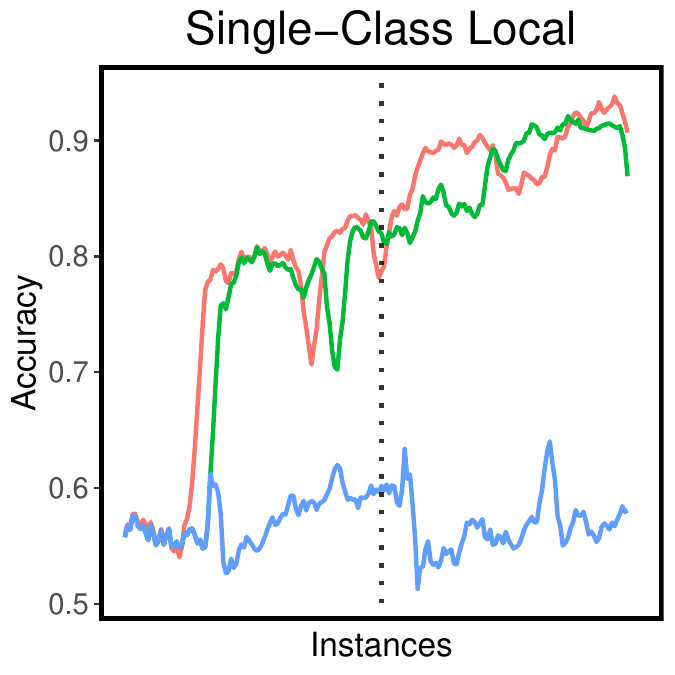}
    \includegraphics[width=.24\textwidth]{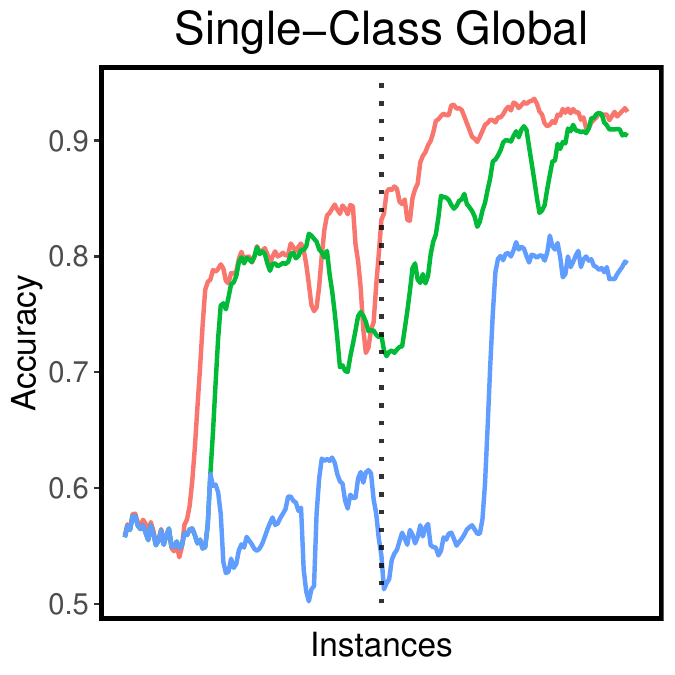}
    \includegraphics[width=.24\textwidth]{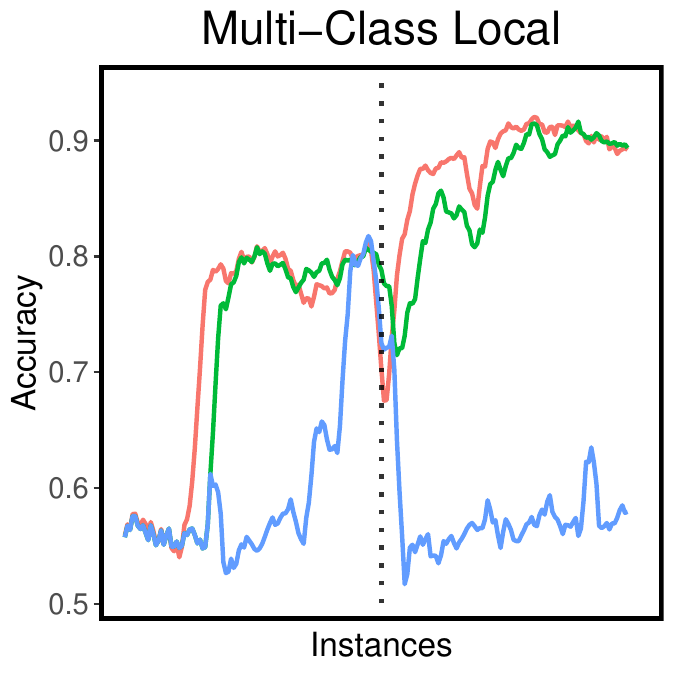}
    \includegraphics[width=.24\textwidth]{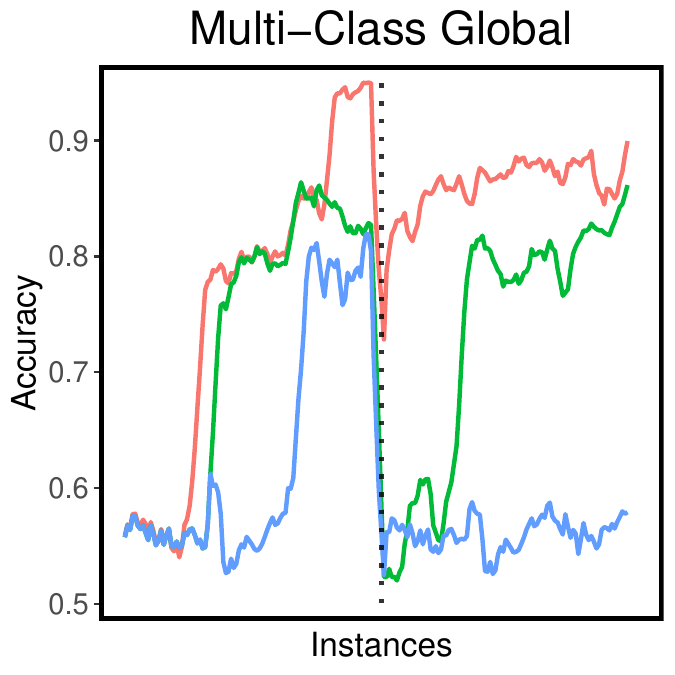}
    \caption{Accuracy curve for AHT, HT and HT-DW on \texttt{reappearing\_cluster} data stream in the presence of a sudden concept drift. Dashed vertical line indicates where the concept drift happened.}
    \label{fig:acc_prune_overall_2}
\end{figure}

\begin{figure}[t!]
    \centering
    \includegraphics[width=\textwidth]{legend.pdf}
    \includegraphics[width=.24\textwidth]{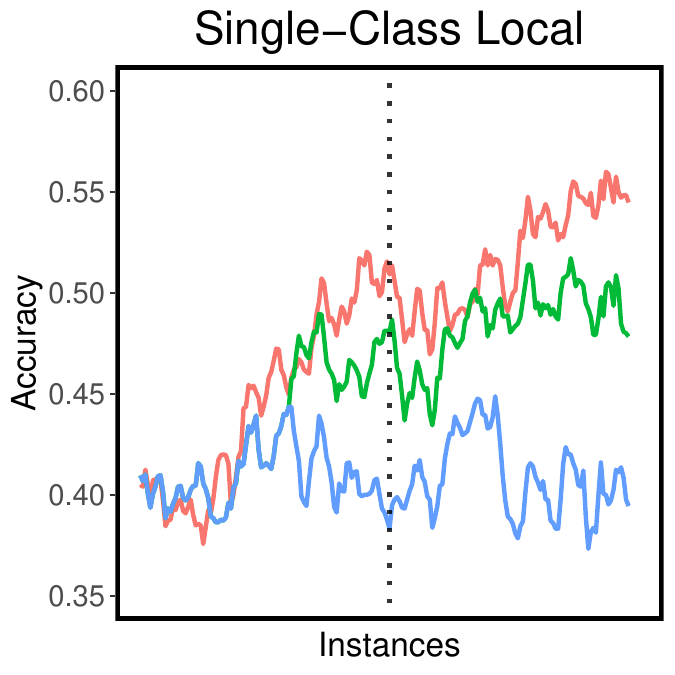}
    \includegraphics[width=.24\textwidth]{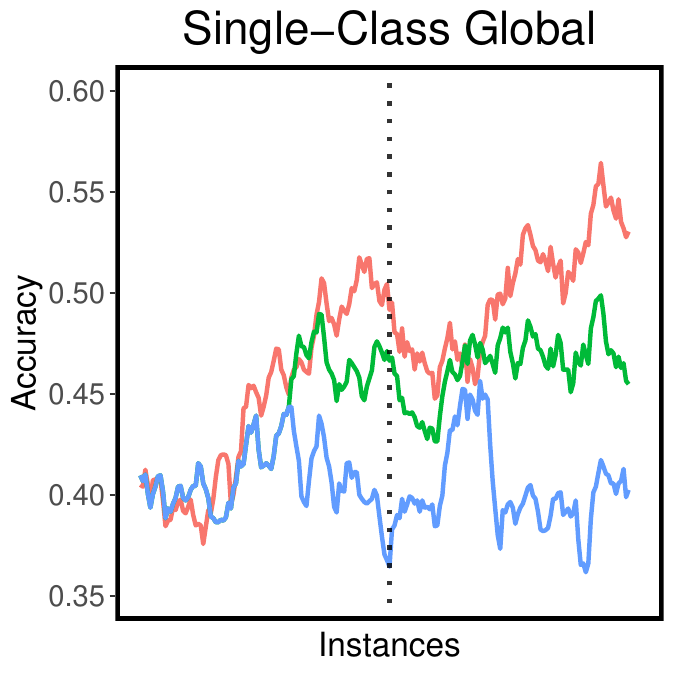}
    \includegraphics[width=.24\textwidth]{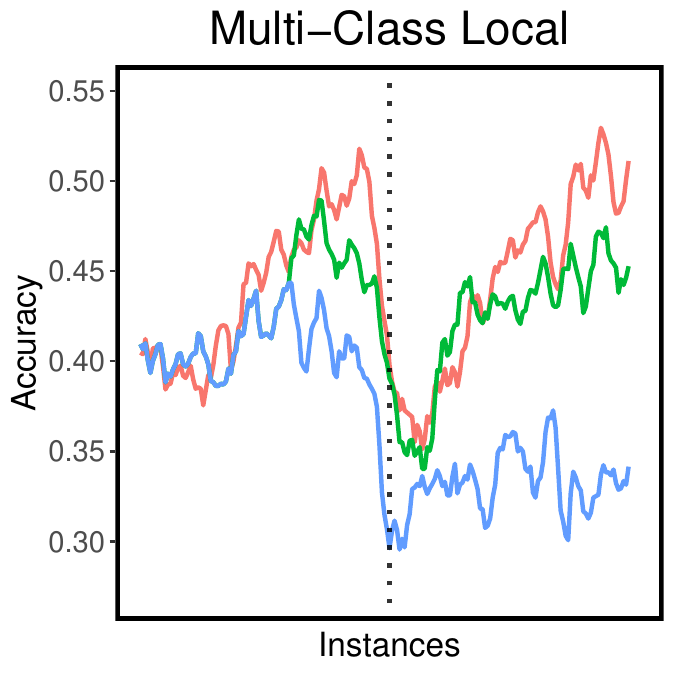}
    \includegraphics[width=.24\textwidth]{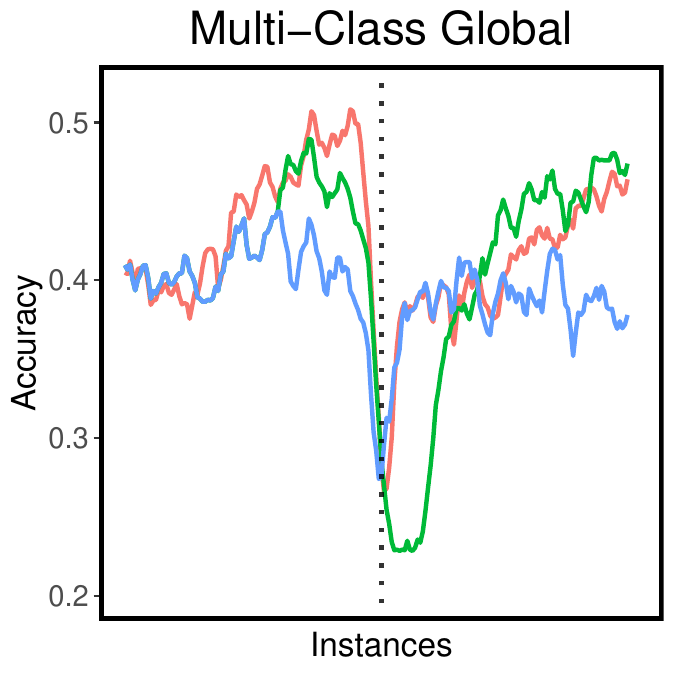}
    \caption{Accuracy curve for AHT, HT and HT-DW on \texttt{prune\_growth\_new\_branch} data stream in the presence of a gradual concept drift. Dashed vertical line indicates where the concept drift happened.}
    \label{fig:acc_prune_overall_3}
\end{figure}

\begin{figure}[t!]
    \centering
    \includegraphics[width=\textwidth]{legend.pdf}
    \includegraphics[width=.24\textwidth]{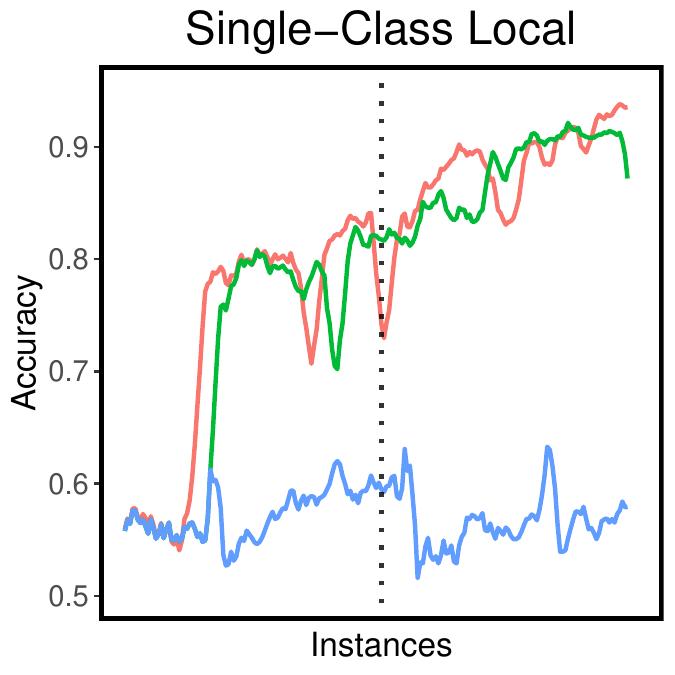}
    \includegraphics[width=.24\textwidth]{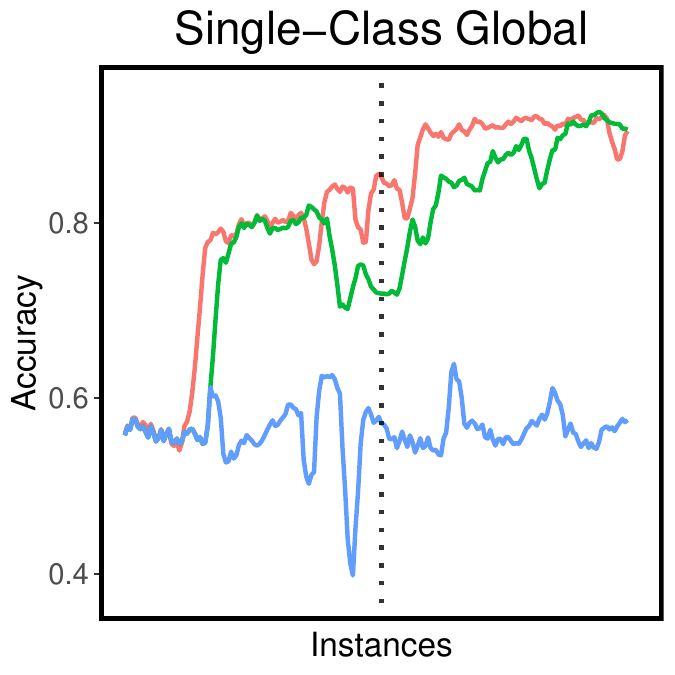}
    \includegraphics[width=.24\textwidth]{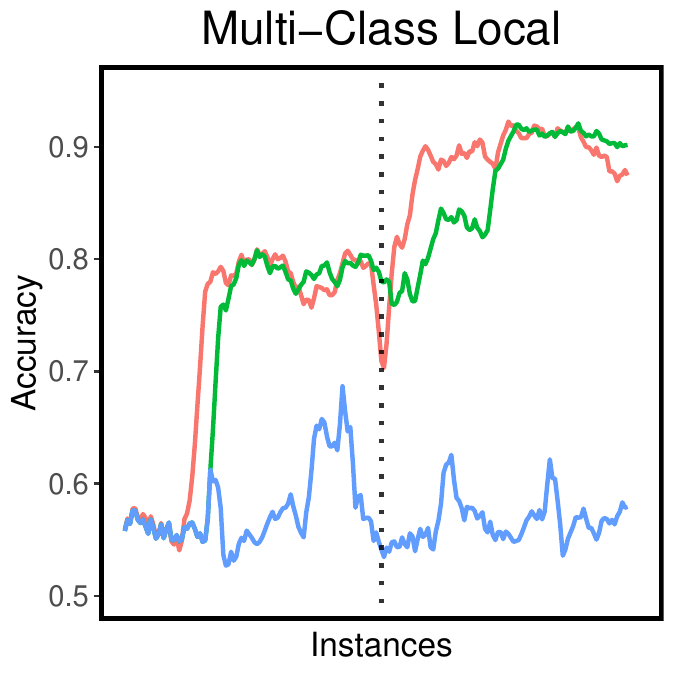}
    \includegraphics[width=.24\textwidth]{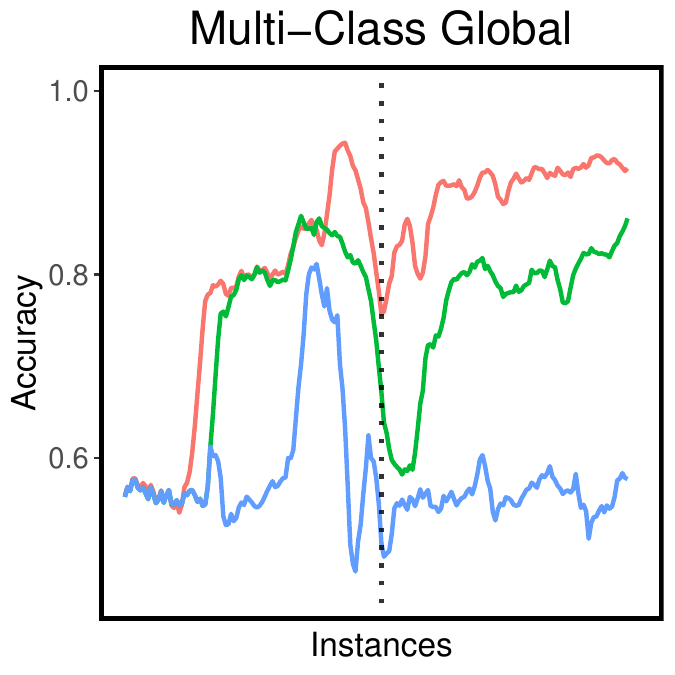}
    \caption{Accuracy curve for AHT, HT and HT-DW on \texttt{reappearing\_cluster} data stream in the presence of a gradual concept drift. Dashed vertical line indicates where the concept drift happened.}
    \label{fig:acc_prune_overall_4}
\end{figure}

Additionally, in Figs.~\ref{fig:acc_prune_overall}-\ref{fig:acc_prune_overall_4}, the accuracy of each classifier under the \texttt{prune\_\-growth\_new\_branch}, \texttt{reappearing\-\_cluster}, \texttt{reappearing\_cluster} and \texttt{reappearing\_cluster} difficulties is displayed, including sudden and gradual drifts. These figures serve as an illustration of the classifiers' behavior across different drift categories. Across Single-Class drifts, both local and global, AHT consistently demonstrated superior accuracy. Even in the presence of drift, HT showcased a remarkable self-adaptation ability, recovering swiftly. Conversely, the retraining strategy resulted in notably poorer performance. Notably, the sole scenario where retraining displayed an advantage was in the context of Multi-Class Global drifts. This specific category provoked a significant change in data distribution, evident from the notable accuracy drop. Given the entirely new data distribution, retraining the classifier proved a valid method for recovering from these drifts. However, by the end of the stream, the plain HT exhibited better accuracy compared to both drift-adapted classifiers.



\begin{table}[h!]
\centering
\caption{Accuracy of HT and AHT for evaluated drift categories.}
\label{tab:acc_drift}
\resizebox{.85\textwidth}{!}{%
\begin{tabular}{@{}lccc@{}}
\toprule
\textbf{Drift Category} & \textbf{HT} & \textbf{AHT} & \textbf{HT-DW} \\ \midrule
Single-Class Local & 66.88\% $\pm$ 0.148 & 70.71\% $\pm$ 0.127 ($\uparrow$ 3.83\%) & 50.13\% $\pm$ 0.135 ($\downarrow$ 16.75\%) \\
Single-Class Global & 68.16\% $\pm$ 0.148 & 71.61\% $\pm$ 0.132 ($\uparrow$ 3.44\%) & 52.55\% $\pm$ 0.157 ($\downarrow$ 15.61\%) \\
Multi-Class Local & 60.78\% $\pm$ 0.147 & 65.18\% $\pm$ 0.126 ($\uparrow$ 4.39\%) & 44.92\% $\pm$ 0.121 ($\downarrow$ 15.87\%)\\
Multi-Class Global & 58.80\% $\pm$ 0.144 & 63.28\% $\pm$ 0.128 ($\uparrow$ 4.48\%)& 45.52\% $\pm$ 0.122 ($\downarrow$ 13.28\%)\\ \bottomrule
\end{tabular}%
}
\end{table}

\clearpage
\section{Lessons learned and recommendations}
\label{sec:lessons}

In order to summarize the knowledge we extracted through the experimental evaluation, this section presents the lessons learned and recommendations for future researchers.

\noindent \textbf{Locality matters.}  The locality of concept drift plays a pivotal role in the accuracy and adaptability of detectors when detecting changes in data streams. When dealing with the detection of concept drift, understanding its locality provides insights into how changes within the data stream, affecting either the entire stream or specific parts of it, influence the classification system. For instance, in scenarios where drift affects only particular sections, such as individual clusters or subsets of classes, it is critical to discern the precise areas within the data stream that undergo these changes. Different sections might require varied adaptations to maintain predictive performance. Moreover, different types of drift, local or global, might require different strategies to handle them effectively. Understanding these subtleties can significantly impact the accuracy and performance of the drift detection and adaptation methods used in learning models.

\noindent \textbf{Handling high number of classes.} The learning task becomes more challenging for both classifiers and drift detectors as the number of classes increases. Our study indicated that when the impact is localized due to fewer classes being affected, the detection of the change becomes more difficult. Hence, drift detectors capable of handling a larger number of classes play a crucial role in effective drift detection. Furthermore, future research should aim at detecting drifts within individual classes, a strategy that could facilitate the adaptation of classifiers to new concepts. 

\noindent \textbf{False alarms.} Drift detectors that rely on error rates often generate numerous false alarms. Given their univariate nature, even positive changes in the error rate, such as when a new branch is created in an incremental tree or a classifier establishes a new boundary, can trigger alarms in certain drift detectors, which is counterproductive. In more complex scenarios, such as localized changes, false alarms can mislead the classifier, resulting in poorer performance. For future research, it is essential that drift detectors strike a better balance between detection and false alarms by examining not only the error curve but also statistics related to data distribution.

The common strategy to address concept drift involves replacing the classifier once a drift is identified. While this approach works well with ensembles due to their varied classifiers, our experiments have shown that when handling drifts that only affect specific boundaries or when the classifier is capable of self-adaptation, avoiding complete replacement of the established boundaries results in better performance. Furthermore, retraining the classifier for each identified drift may lead to redundant work due to the high occurrence of false alarms, even with state-of-the-art drift detectors. This emphasizes the necessity for drift detectors capable of pinpointing which class or segment of the data stream was affected by the drift. This, in turn, facilitates a more precise adaptation of the classifier rather than a complete replacement.

\noindent \textbf{Retraining is not (always) the best option.} The common strategy to address concept drift involves replacing the classifier once a drift is identified. While this approach works well with ensembles due to their varied classifiers, our experiments have shown that when handling drifts that only affect specific boundaries or when the classifier is capable of self-adaptation, avoiding complete replacement of the established boundaries results in better performance. Furthermore, retraining the classifier for each identified drift may lead to redundant work due to the high occurrence of false alarms, even with state-of-the-art drift detectors. This emphasizes the necessity for drift detectors capable of pinpointing which class or segment of the data stream was affected by the drift. This, in turn, facilitates a more precise adaptation of the classifier rather than a complete replacement.

\section{Conclusion and future work}
\label{sec:conclusions}

In this paper, we presented a comprehensive study focusing on benchmarking and evaluating the impact of concept drift, specifically concerning its locality and magnitude, on classifiers and drift detectors. We introduced a novel categorization of concept drift, considering its locality and scale. Through a systematic approach, we identified significant challenges within this domain and formulated a set of $2,760$ benchmark problems that encompass various levels of difficulty, guided by our proposed categorization. Additionally, we conducted a comparative evaluation of $9$ state-of-the-art drift detectors across a diverse range of difficulties. Each drift detector was thoroughly assessed across various isolated and combined scenarios. This analysis not only identifies the top-performing detectors but also sheds light on their specific strengths, providing valuable insights for future research in drift detection. Moreover, we evaluated how the locality of drift influences the base-classifier's performance, gaining knowledge into the most effective approaches for addressing different categories of concept drift to minimize recovery time. This comprehensive study offered insights into different and yet unexplored categorizations of concept drift, providing an understanding of how drift detectors and classifiers perform across numerous difficulties. All benchmark problems and evaluation methodologies are publicly available, enabling future researchers to design efficient approaches, drift detectors, and classifiers to handle local concept drifts effectively.

Our future work aims to explore the influence of concept drift locality on data streams characterized by imbalanced class distributions. Additionally, our plans include an investigation into the performance of unsupervised and semi-supervised drift detectors within the proposed scenarios. Furthermore, we seek to develop classifiers with the capacity to handle concept drift locally, avoiding the necessity for retraining or complete replacement.


\bibliographystyle{elsarticle-num-names} 
\bibliography{references}

\begin{thebibliography}{50}
\expandafter\ifx\csname natexlab\endcsname\relax\def\natexlab#1{#1}\fi
\providecommand{\url}[1]{\texttt{#1}}
\providecommand{\href}[2]{#2}
\providecommand{\path}[1]{#1}
\providecommand{\DOIprefix}{doi:}
\providecommand{\ArXivprefix}{arXiv:}
\providecommand{\URLprefix}{URL: }
\providecommand{\Pubmedprefix}{pmid:}
\providecommand{\doi}[1]{\href{http://dx.doi.org/#1}{\path{#1}}}
\providecommand{\Pubmed}[1]{\href{pmid:#1}{\path{#1}}}
\providecommand{\bibinfo}[2]{#2}
\ifx\xfnm\relax \def\xfnm[#1]{\unskip,\space#1}\fi
\bibitem[{Gama et~al.(2004)Gama, Medas, Castillo, and Rodrigues}]{gama2004learning}
\bibinfo{author}{J.~Gama}, \bibinfo{author}{P.~Medas}, \bibinfo{author}{G.~Castillo}, \bibinfo{author}{P.~Rodrigues},
\newblock \bibinfo{title}{Learning with drift detection},
\newblock in: \bibinfo{booktitle}{Brazilian Symposium on Artificial Intelligence}, \bibinfo{year}{2004}.
\bibitem[{Bahri et~al.(2021)Bahri, Bifet, Gama, Gomes, and Maniu}]{bahri2021data}
\bibinfo{author}{M.~Bahri}, \bibinfo{author}{A.~Bifet}, \bibinfo{author}{J.~Gama}, \bibinfo{author}{H.~M. Gomes}, \bibinfo{author}{S.~Maniu},
\newblock \bibinfo{title}{Data stream analysis: Foundations, major tasks and tools},
\newblock \bibinfo{journal}{Wiley Interdisciplinary Reviews: Data Mining and Knowledge Discovery}  (\bibinfo{year}{2021}).
\bibitem[{Gama(2010)}]{gama2010knowledge}
\bibinfo{author}{J.~Gama}, \bibinfo{title}{Knowledge discovery from data streams}, \bibinfo{publisher}{CRC Press}, \bibinfo{year}{2010}.
\bibitem[{Gama et~al.(2014)Gama, Žliobaitė, Bifet, Pechenizkiy, and Bouchachia}]{gama2014survey}
\bibinfo{author}{J.~Gama}, \bibinfo{author}{I.~Žliobaitė}, \bibinfo{author}{A.~Bifet}, \bibinfo{author}{M.~Pechenizkiy}, \bibinfo{author}{A.~Bouchachia},
\newblock \bibinfo{title}{A survey on concept drift adaptation},
\newblock \bibinfo{journal}{ACM Computing Surveys}  (\bibinfo{year}{2014}).
\bibitem[{Aguiar et~al.(2023)Aguiar, Krawczyk, and Cano}]{aguiar2022survey}
\bibinfo{author}{G.~Aguiar}, \bibinfo{author}{B.~Krawczyk}, \bibinfo{author}{A.~Cano},
\newblock \bibinfo{title}{A survey on learning from imbalanced data streams: taxonomy, challenges, empirical study, and reproducible experimental framework},
\newblock \bibinfo{journal}{Machine Learning}  (\bibinfo{year}{2023}).
\bibitem[{Korycki and Krawczyk(2021)}]{korycki2021concept}
\bibinfo{author}{{\L}.~Korycki}, \bibinfo{author}{B.~Krawczyk},
\newblock \bibinfo{title}{Concept drift detection from multi-class imbalanced data streams},
\newblock in: \bibinfo{booktitle}{IEEE International Conference on Data Engineering (ICDE)}, \bibinfo{year}{2021}.
\bibitem[{Viniski et~al.(2021)Viniski, Barddal, de~Souza Britto~Jr, Enembreck, and de~Campos}]{viniski2021case}
\bibinfo{author}{A.~D. Viniski}, \bibinfo{author}{J.~P. Barddal}, \bibinfo{author}{A.~de~Souza Britto~Jr}, \bibinfo{author}{F.~Enembreck}, \bibinfo{author}{H.~V.~A. de~Campos},
\newblock \bibinfo{title}{A case study of batch and incremental recommender systems in supermarket data under concept drifts and cold start},
\newblock \bibinfo{journal}{Expert Systems with Applications}  (\bibinfo{year}{2021}).
\bibitem[{Su{\'a}rez-Cetrulo et~al.(2023)Su{\'a}rez-Cetrulo, Quintana, and Cervantes}]{suarez2022survey}
\bibinfo{author}{A.~L. Su{\'a}rez-Cetrulo}, \bibinfo{author}{D.~Quintana}, \bibinfo{author}{A.~Cervantes},
\newblock \bibinfo{title}{A survey on machine learning for recurring concept drifting data streams},
\newblock \bibinfo{journal}{Expert Systems with Applications}  (\bibinfo{year}{2023}).
\bibitem[{Lu et~al.(2016)Lu, Lu, Zhang, and De~Mantaras}]{lu2016concept}
\bibinfo{author}{N.~Lu}, \bibinfo{author}{J.~Lu}, \bibinfo{author}{G.~Zhang}, \bibinfo{author}{R.~L. De~Mantaras},
\newblock \bibinfo{title}{A concept drift-tolerant case-base editing technique},
\newblock \bibinfo{journal}{Artificial Intelligence}  (\bibinfo{year}{2016}).
\bibitem[{Liu et~al.(2017)Liu, Song, Zhang, and Lu}]{liu2017regional}
\bibinfo{author}{A.~Liu}, \bibinfo{author}{Y.~Song}, \bibinfo{author}{G.~Zhang}, \bibinfo{author}{J.~Lu},
\newblock \bibinfo{title}{Regional concept drift detection and density synchronized drift adaptation},
\newblock in: \bibinfo{booktitle}{International Joint Conference on Artificial Intelligence (IJCAI)}, \bibinfo{year}{2017}.
\bibitem[{Barros et~al.(2017)Barros, Cabral, Gon{\c{c}}alves~Jr, and Santos}]{barros2017rddm}
\bibinfo{author}{R.~S. Barros}, \bibinfo{author}{D.~R. Cabral}, \bibinfo{author}{P.~M. Gon{\c{c}}alves~Jr}, \bibinfo{author}{S.~G. Santos},
\newblock \bibinfo{title}{Rddm: Reactive drift detection method},
\newblock \bibinfo{journal}{Expert Systems with Applications}  (\bibinfo{year}{2017}).
\bibitem[{Gulcan and Can(2023)}]{gulcan2023unsupervised}
\bibinfo{author}{E.~B. Gulcan}, \bibinfo{author}{F.~Can},
\newblock \bibinfo{title}{Unsupervised concept drift detection for multi-label data streams},
\newblock \bibinfo{journal}{Artificial Intelligence Review}  (\bibinfo{year}{2023}).
\bibitem[{Gama and Castillo(2006)}]{gama2006learning}
\bibinfo{author}{J.~Gama}, \bibinfo{author}{G.~Castillo},
\newblock \bibinfo{title}{Learning with local drift detection},
\newblock in: \bibinfo{booktitle}{International conference on advanced data mining and applications}, \bibinfo{organization}{Springer}, \bibinfo{year}{2006}.
\bibitem[{Krawczyk et~al.(2017)Krawczyk, Minku, Gama, Stefanowski, and Wo{\'z}niak}]{krawczyk2017ensemble}
\bibinfo{author}{B.~Krawczyk}, \bibinfo{author}{L.~L. Minku}, \bibinfo{author}{J.~Gama}, \bibinfo{author}{J.~Stefanowski}, \bibinfo{author}{M.~Wo{\'z}niak},
\newblock \bibinfo{title}{Ensemble learning for data stream analysis: A survey},
\newblock \bibinfo{journal}{Information Fusion}  (\bibinfo{year}{2017}).
\bibitem[{Lu et~al.(2018)Lu, Liu, Dong, Gu, Gama, and Zhang}]{lu2018learning}
\bibinfo{author}{J.~Lu}, \bibinfo{author}{A.~Liu}, \bibinfo{author}{F.~Dong}, \bibinfo{author}{F.~Gu}, \bibinfo{author}{J.~Gama}, \bibinfo{author}{G.~Zhang},
\newblock \bibinfo{title}{Learning under concept drift: A review},
\newblock \bibinfo{journal}{IEEE Transactions on Knowledge and Data Engineering}  (\bibinfo{year}{2018}).
\bibitem[{Masegosa et~al.(2020)Masegosa, Mart{\'\i}nez, Ramos-L{\'o}pez, Langseth, Nielsen, and Salmer{\'o}n}]{masegosa2020analyzing}
\bibinfo{author}{A.~R. Masegosa}, \bibinfo{author}{A.~M. Mart{\'\i}nez}, \bibinfo{author}{D.~Ramos-L{\'o}pez}, \bibinfo{author}{H.~Langseth}, \bibinfo{author}{T.~D. Nielsen}, \bibinfo{author}{A.~Salmer{\'o}n},
\newblock \bibinfo{title}{Analyzing concept drift: A case study in the financial sector},
\newblock \bibinfo{journal}{Intelligent Data Analysis}  (\bibinfo{year}{2020}).
\bibitem[{Webb et~al.(2016)Webb, Hyde, Cao, Nguyen, and Petitjean}]{webb2016characterizing}
\bibinfo{author}{G.~I. Webb}, \bibinfo{author}{R.~Hyde}, \bibinfo{author}{H.~Cao}, \bibinfo{author}{H.~L. Nguyen}, \bibinfo{author}{F.~Petitjean},
\newblock \bibinfo{title}{Characterizing concept drift},
\newblock \bibinfo{journal}{Data Mining and Knowledge Discovery}  (\bibinfo{year}{2016}).
\bibitem[{Krawczyk and Cano(2018)}]{krawczyk2018online}
\bibinfo{author}{B.~Krawczyk}, \bibinfo{author}{A.~Cano},
\newblock \bibinfo{title}{Online ensemble learning with abstaining classifiers for drifting and noisy data streams},
\newblock \bibinfo{journal}{Applied Soft Computing}  (\bibinfo{year}{2018}).
\bibitem[{Ditzler et~al.(2015)Ditzler, Roveri, Alippi, and Polikar}]{ditzler2015learning}
\bibinfo{author}{G.~Ditzler}, \bibinfo{author}{M.~Roveri}, \bibinfo{author}{C.~Alippi}, \bibinfo{author}{R.~Polikar},
\newblock \bibinfo{title}{Learning in nonstationary environments: A survey},
\newblock \bibinfo{journal}{IEEE Computational Intelligence Magazine}  (\bibinfo{year}{2015}).
\bibitem[{Barros and Santos(2018)}]{barros2018large}
\bibinfo{author}{R.~S.~M. Barros}, \bibinfo{author}{S.~G. T.~C. Santos},
\newblock \bibinfo{title}{A large-scale comparison of concept drift detectors},
\newblock \bibinfo{journal}{Information Sciences}  (\bibinfo{year}{2018}).
\bibitem[{Page(1954)}]{page1954continuous}
\bibinfo{author}{E.~S. Page},
\newblock \bibinfo{title}{Continuous inspection schemes},
\newblock \bibinfo{journal}{Biometrika}  (\bibinfo{year}{1954}).
\bibitem[{Roberts(2000)}]{roberts2000control}
\bibinfo{author}{S.~Roberts},
\newblock \bibinfo{title}{Control chart tests based on geometric moving averages},
\newblock \bibinfo{journal}{Technometrics}  (\bibinfo{year}{2000}).
\bibitem[{Baena-Garc{\i}a et~al.(2006)Baena-Garc{\i}a, del Campo-{\'A}vila, Fidalgo, Bifet, Gavalda, and Morales-Bueno}]{baena2006early}
\bibinfo{author}{M.~Baena-Garc{\i}a}, \bibinfo{author}{J.~del Campo-{\'A}vila}, \bibinfo{author}{R.~Fidalgo}, \bibinfo{author}{A.~Bifet}, \bibinfo{author}{R.~Gavalda}, \bibinfo{author}{R.~Morales-Bueno},
\newblock \bibinfo{title}{Early drift detection method},
\newblock in: \bibinfo{booktitle}{International Workshop on Knowledge Discovery from Data Streams}, \bibinfo{year}{2006}.
\bibitem[{Frias-Blanco et~al.(2014)Frias-Blanco, del Campo-{\'A}vila, Ramos-Jimenez, Morales-Bueno, Ortiz-D{\'\i}az, and Caballero-Mota}]{frias2014online}
\bibinfo{author}{I.~Frias-Blanco}, \bibinfo{author}{J.~del Campo-{\'A}vila}, \bibinfo{author}{G.~Ramos-Jimenez}, \bibinfo{author}{R.~Morales-Bueno}, \bibinfo{author}{A.~Ortiz-D{\'\i}az}, \bibinfo{author}{Y.~Caballero-Mota},
\newblock \bibinfo{title}{Online and non-parametric drift detection methods based on hoeffding’s bounds},
\newblock \bibinfo{journal}{IEEE Transactions on Knowledge and Data Engineering}  (\bibinfo{year}{2014}).
\bibitem[{de~Barros et~al.(2018)de~Barros, Hidalgo, and de~Lima~Cabral}]{de2018wilcoxon}
\bibinfo{author}{R.~S.~M. de~Barros}, \bibinfo{author}{J.~I.~G. Hidalgo}, \bibinfo{author}{D.~R. de~Lima~Cabral},
\newblock \bibinfo{title}{Wilcoxon rank sum test drift detector},
\newblock \bibinfo{journal}{Neurocomputing}  (\bibinfo{year}{2018}).
\bibitem[{Ross et~al.(2012)Ross, Adams, Tasoulis, and Hand}]{ross2012exponentially}
\bibinfo{author}{G.~J. Ross}, \bibinfo{author}{N.~M. Adams}, \bibinfo{author}{D.~K. Tasoulis}, \bibinfo{author}{D.~J. Hand},
\newblock \bibinfo{title}{Exponentially weighted moving average charts for detecting concept drift},
\newblock \bibinfo{journal}{Pattern Recognition Letters}  (\bibinfo{year}{2012}).
\bibitem[{Bifet and Gavalda(2007)}]{bifet2007learning}
\bibinfo{author}{A.~Bifet}, \bibinfo{author}{R.~Gavalda},
\newblock \bibinfo{title}{Learning from time-changing data with adaptive windowing},
\newblock in: \bibinfo{booktitle}{SIAM International Conference on Data Mining}, \bibinfo{year}{2007}.
\bibitem[{Raab et~al.(2020)Raab, Heusinger, and Schleif}]{raab2020reactive}
\bibinfo{author}{C.~Raab}, \bibinfo{author}{M.~Heusinger}, \bibinfo{author}{F.-M. Schleif},
\newblock \bibinfo{title}{Reactive soft prototype computing for concept drift streams},
\newblock \bibinfo{journal}{Neurocomputing}  (\bibinfo{year}{2020}).
\bibitem[{Nishida and Yamauchi(2007)}]{nishida2007detecting}
\bibinfo{author}{K.~Nishida}, \bibinfo{author}{K.~Yamauchi},
\newblock \bibinfo{title}{Detecting concept drift using statistical testing},
\newblock in: \bibinfo{booktitle}{International Conference on Discovery Science}, \bibinfo{organization}{Springer}, \bibinfo{year}{2007}.
\bibitem[{Huang et~al.(2014)Huang, Koh, Dobbie, and Pears}]{huang2014detecting}
\bibinfo{author}{D.~T.~J. Huang}, \bibinfo{author}{Y.~S. Koh}, \bibinfo{author}{G.~Dobbie}, \bibinfo{author}{R.~Pears},
\newblock \bibinfo{title}{Detecting volatility shift in data streams},
\newblock in: \bibinfo{booktitle}{IEEE International Conference on Data Mining}, \bibinfo{year}{2014}.
\bibitem[{Pesaranghader and Viktor(2016)}]{pesaranghader2016fast}
\bibinfo{author}{A.~Pesaranghader}, \bibinfo{author}{H.~Viktor},
\newblock \bibinfo{title}{Fast hoeffding drift detection method for evolving data streams},
\newblock in: \bibinfo{booktitle}{Machine Learning and Knowledge Discovery in Databases: European Conference}, \bibinfo{year}{2016}.
\bibitem[{Komorniczak and Ksieniewicz(2023)}]{komorniczak2023complexity}
\bibinfo{author}{J.~Komorniczak}, \bibinfo{author}{P.~Ksieniewicz},
\newblock \bibinfo{title}{Complexity-based drift detection for nonstationary data streams},
\newblock \bibinfo{journal}{Neurocomputing}  (\bibinfo{year}{2023}).
\bibitem[{Wang et~al.(2024)Wang, Yu, Jin, Davies, and Woo}]{wang2024quadcdd}
\bibinfo{author}{P.~Wang}, \bibinfo{author}{H.~Yu}, \bibinfo{author}{N.~Jin}, \bibinfo{author}{D.~Davies}, \bibinfo{author}{W.~L. Woo},
\newblock \bibinfo{title}{Quadcdd: A quadruple-based approach for understanding concept drift in data streams},
\newblock \bibinfo{journal}{Expert Systems with Applications}  (\bibinfo{year}{2024}).
\bibitem[{{\L}api{\'n}ski et~al.(2018){\L}api{\'n}ski, Krawczyk, Ksicnicwicz, and Wo{\'z}niak}]{lapinski2018empirical}
\bibinfo{author}{A.~{\L}api{\'n}ski}, \bibinfo{author}{B.~Krawczyk}, \bibinfo{author}{P.~Ksicnicwicz}, \bibinfo{author}{M.~Wo{\'z}niak},
\newblock \bibinfo{title}{An empirical insight into concept drift detectors ensemble strategies},
\newblock in: \bibinfo{booktitle}{IEEE Congress on Evolutionary Computation}, \bibinfo{year}{2018}.
\bibitem[{Sobolewski and Wo{\'z}niak(2013)}]{sobolewski2013comparable}
\bibinfo{author}{P.~Sobolewski}, \bibinfo{author}{M.~Wo{\'z}niak},
\newblock \bibinfo{title}{Comparable study of statistical tests for virtual concept drift detection},
\newblock in: \bibinfo{booktitle}{International Conference on Computer Recognition Systems}, \bibinfo{year}{2013}.
\bibitem[{Song et~al.(2007)Song, Wu, Jermaine, and Ranka}]{song2007statistical}
\bibinfo{author}{X.~Song}, \bibinfo{author}{M.~Wu}, \bibinfo{author}{C.~Jermaine}, \bibinfo{author}{S.~Ranka},
\newblock \bibinfo{title}{Statistical change detection for multi-dimensional data},
\newblock in: \bibinfo{booktitle}{ACM SIGKDD International Conference on Knowledge Discovery and Data Mining}, \bibinfo{year}{2007}.
\bibitem[{Qahtan et~al.(2015)Qahtan, Alharbi, Wang, and Zhang}]{qahtan2015pca}
\bibinfo{author}{A.~A. Qahtan}, \bibinfo{author}{B.~Alharbi}, \bibinfo{author}{S.~Wang}, \bibinfo{author}{X.~Zhang},
\newblock \bibinfo{title}{A pca-based change detection framework for multidimensional data streams: Change detection in multidimensional data streams},
\newblock in: \bibinfo{booktitle}{ACM SIGKDD International Conference on Knowledge Discovery and Data Mining}, \bibinfo{year}{2015}.
\bibitem[{Gu et~al.(2016)Gu, Zhang, Lu, and Lin}]{gu2016concept}
\bibinfo{author}{F.~Gu}, \bibinfo{author}{G.~Zhang}, \bibinfo{author}{J.~Lu}, \bibinfo{author}{C.-T. Lin},
\newblock \bibinfo{title}{Concept drift detection based on equal density estimation},
\newblock in: \bibinfo{booktitle}{International Joint Conference on Neural Networks (IJCNN)}, \bibinfo{organization}{IEEE}, \bibinfo{year}{2016}.
\bibitem[{Bu et~al.(2016)Bu, Alippi, and Zhao}]{bu2016pdf}
\bibinfo{author}{L.~Bu}, \bibinfo{author}{C.~Alippi}, \bibinfo{author}{D.~Zhao},
\newblock \bibinfo{title}{A pdf-free change detection test based on density difference estimation},
\newblock \bibinfo{journal}{IEEE Transactions on Neural Networks and Learning Systems}  (\bibinfo{year}{2016}).
\bibitem[{Liu et~al.(2018)Liu, Lu, Liu, and Zhang}]{liu2018accumulating}
\bibinfo{author}{A.~Liu}, \bibinfo{author}{J.~Lu}, \bibinfo{author}{F.~Liu}, \bibinfo{author}{G.~Zhang},
\newblock \bibinfo{title}{Accumulating regional density dissimilarity for concept drift detection in data streams},
\newblock \bibinfo{journal}{Pattern Recognition}  (\bibinfo{year}{2018}).
\bibitem[{Lu et~al.(2014)Lu, Zhang, and Lu}]{lu2014concept}
\bibinfo{author}{N.~Lu}, \bibinfo{author}{G.~Zhang}, \bibinfo{author}{J.~Lu},
\newblock \bibinfo{title}{Concept drift detection via competence models},
\newblock \bibinfo{journal}{Artificial Intelligence}  (\bibinfo{year}{2014}).
\bibitem[{Gon{\c{c}}alves~Jr et~al.(2014)Gon{\c{c}}alves~Jr, de~Carvalho~Santos, Barros, and Vieira}]{gonccalves2014comparative}
\bibinfo{author}{P.~M. Gon{\c{c}}alves~Jr}, \bibinfo{author}{S.~G. de~Carvalho~Santos}, \bibinfo{author}{R.~S. Barros}, \bibinfo{author}{D.~C. Vieira},
\newblock \bibinfo{title}{A comparative study on concept drift detectors},
\newblock \bibinfo{journal}{Expert Systems with Applications}  (\bibinfo{year}{2014}).
\bibitem[{Santos et~al.(2019)Santos, Barros, and Gon{\c{c}}alves~Jr}]{santos2019differential}
\bibinfo{author}{S.~G. Santos}, \bibinfo{author}{R.~S. Barros}, \bibinfo{author}{P.~M. Gon{\c{c}}alves~Jr},
\newblock \bibinfo{title}{A differential evolution based method for tuning concept drift detectors in data streams},
\newblock \bibinfo{journal}{Information Sciences}  (\bibinfo{year}{2019}).
\bibitem[{Bab{\"u}ro{\u{g}}lu et~al.(2021)Bab{\"u}ro{\u{g}}lu, Durmu{\c{s}}o{\u{g}}lu, and Dereli}]{baburouglu2021novel}
\bibinfo{author}{E.~S. Bab{\"u}ro{\u{g}}lu}, \bibinfo{author}{A.~Durmu{\c{s}}o{\u{g}}lu}, \bibinfo{author}{T.~Dereli},
\newblock \bibinfo{title}{Novel hybrid pair recommendations based on a large-scale comparative study of concept drift detection},
\newblock \bibinfo{journal}{Expert Systems with Applications}  (\bibinfo{year}{2021}).
\bibitem[{Poenaru-Olaru et~al.(2022)Poenaru-Olaru, Cruz, van Deursen, and Rellermeyer}]{poenaru2022concept}
\bibinfo{author}{L.~Poenaru-Olaru}, \bibinfo{author}{L.~Cruz}, \bibinfo{author}{A.~van Deursen}, \bibinfo{author}{J.~S. Rellermeyer},
\newblock \bibinfo{title}{Are concept drift detectors reliable alarming systems?-a comparative study},
\newblock in: \bibinfo{booktitle}{IEEE International Conference on Big Data}, \bibinfo{organization}{IEEE}, \bibinfo{year}{2022}.
\bibitem[{Sakurai et~al.(2023)Sakurai, Lopes, Zarpel{\~a}o, and Barbon~Junior}]{sakurai2023benchmarking}
\bibinfo{author}{G.~Y. Sakurai}, \bibinfo{author}{J.~F. Lopes}, \bibinfo{author}{B.~B. Zarpel{\~a}o}, \bibinfo{author}{S.~Barbon~Junior},
\newblock \bibinfo{title}{Benchmarking change detector algorithms from different concept drift perspectives},
\newblock \bibinfo{journal}{Future Internet}  (\bibinfo{year}{2023}).
\bibitem[{Brzezinski et~al.(2021)Brzezinski, Minku, Pewinski, Stefanowski, and Szumaczuk}]{brzezinski2021impact}
\bibinfo{author}{D.~Brzezinski}, \bibinfo{author}{L.~L. Minku}, \bibinfo{author}{T.~Pewinski}, \bibinfo{author}{J.~Stefanowski}, \bibinfo{author}{A.~Szumaczuk},
\newblock \bibinfo{title}{The impact of data difficulty factors on classification of imbalanced and concept drifting data streams},
\newblock \bibinfo{journal}{Knowledge and Information Systems}  (\bibinfo{year}{2021}).
\bibitem[{Lango and Stefanowski(2022)}]{lango2022makes}
\bibinfo{author}{M.~Lango}, \bibinfo{author}{J.~Stefanowski},
\newblock \bibinfo{title}{What makes multi-class imbalanced problems difficult? an experimental study},
\newblock \bibinfo{journal}{Expert Systems with Applications}  (\bibinfo{year}{2022}).
\bibitem[{Holmes et~al.(2005)Holmes, Kirkby, and Pfahringer}]{holmes2005stress}
\bibinfo{author}{G.~Holmes}, \bibinfo{author}{R.~Kirkby}, \bibinfo{author}{B.~Pfahringer},
\newblock \bibinfo{title}{Stress-testing hoeffding trees},
\newblock in: \bibinfo{booktitle}{Knowledge Discovery in Databases: PKDD 2005: 9th European Conference on Principles and Practice of Knowledge Discovery in Databases}, \bibinfo{organization}{Springer}, \bibinfo{year}{2005}.
\bibitem[{Montiel et~al.(2021)Montiel, Halford, Mastelini, Bolmier, Sourty, Vaysse, Zouitine, Gomes, Read, Abdessalem et~al.}]{montiel2021river}
\bibinfo{author}{J.~Montiel}, \bibinfo{author}{M.~Halford}, \bibinfo{author}{S.~M. Mastelini}, \bibinfo{author}{G.~Bolmier}, \bibinfo{author}{R.~Sourty}, \bibinfo{author}{R.~Vaysse}, \bibinfo{author}{A.~Zouitine}, \bibinfo{author}{H.~M. Gomes}, \bibinfo{author}{J.~Read}, \bibinfo{author}{T.~Abdessalem}, et~al.,
\newblock \bibinfo{title}{River: machine learning for streaming data in python},
\newblock \bibinfo{journal}{The Journal of Machine Learning Research}  (\bibinfo{year}{2021}).

\end{thebibliography}


\end{document}